\documentclass[oupdraft]{bio}
\usepackage{amsmath}
\usepackage{graphicx}
\usepackage{enumerate}
\usepackage{url} 
\usepackage[utf8]{inputenc}
\usepackage{booktabs}
\usepackage{amsthm}
\usepackage{amssymb}
\usepackage[colorinlistoftodos]{todonotes}
\usepackage{caption}
\usepackage{subcaption}
\usepackage{lscape}
\usepackage{verbatim}
\usepackage{csvsimple}
\usepackage{multirow}
\usepackage{textcomp}
\usepackage{natbib}
\usepackage[linesnumbered,ruled,vlined]{algorithm2e}
\usepackage{bm}
\usepackage{wrapfig}
\usepackage{pdfpages,setspace}
\usepackage[normalem]{ulem}

\newcommand{\balpha}{\boldsymbol{\alpha}}
\newcommand{\bdel}{\boldsymbol{\delta}}
\newcommand{\bphi}{\boldsymbol{\phi}}

\newcommand{\btheta}{\boldsymbol{\theta}}
\newcommand{\bzeta}{\boldsymbol{\zeta}}

\newcommand{\wt}{\widetilde}

\begin{document}

\title{HMM for Discovering Decision-Making Dynamics Using Reinforcement Learning Experiments}

\author{XINGCHE GUO$^1$, DONGLIN ZENG$^2$, YUANJIA WANG$^{1,3 \ \ast}$ \\[4pt]
$^1$\textit{Department of Biostatistics, Columbia University, New York, U.S.A.} \\
$^2$\textit{Department of Biostatistics, University of Michigan, Ann Arbor, U.S.A.} \\
$^3$\textit{Department of Psychiatry, Columbia University, New York, U.S.A.} \\[2pt]
{yw2016@cumc.columbia.edu}}

\markboth%
{X.\ Guo, D.\ Zeng and Y.\ Wang}
{HMM for decision-making dynamics using RL experiments}

\maketitle

\footnotetext{Dr.\ Yuanjia Wang is the corresponding author.}

\begin{abstract}
{Major depressive disorder (MDD), a leading cause of years of life lived with disability, presents challenges in diagnosis and treatment due to its complex and heterogeneous nature. Emerging evidence indicates that reward processing abnormalities may serve as a behavioral marker for MDD. 
To measure reward processing, patients perform computer-based behavioral tasks that involve making choices or responding to stimulants that are associated with different outcomes, such as gains or losses in the laboratory. Reinforcement learning (RL) models are fitted to extract parameters that measure various aspects of reward processing (e.g., reward sensitivity) to characterize how patients make decisions in behavioral tasks. Recent findings suggest the inadequacy of characterizing reward learning solely based on a single RL model; instead, there may be a switching of decision-making processes between multiple strategies.
An important scientific question is how the dynamics of strategies in decision-making affect the reward learning ability of individuals with MDD.
Motivated by the probabilistic reward task (PRT) within the EMBARC study, we propose a novel RL-HMM framework for analyzing reward-based decision-making. Our model accommodates decision-making strategy switching between two distinct approaches under a hidden Markov model (HMM): subjects making decisions based on the RL model or opting for random choices. We account for continuous RL state space and allow time-varying transition probabilities in the HMM.
We introduce a computationally efficient EM algorithm for parameter estimation and employ a nonparametric bootstrap for inference. Extensive simulation studies validate the finite-sample performance of our method.
We apply our approach to the EMBARC study to show that MDD patients are less engaged in RL compared to the healthy controls, and engagement is associated with brain activities in the negative affect circuitry during an emotional conflict task.
\vspace{-2em}
}
{reinforcement learning; reward tasks; state-switching; behavioral phenotyping; mental health; brain-behavior association}
\end{abstract}

\vspace{-3em}
\section{Introduction}
\label{sec:intro}

Despite extensive research and clinical efforts dedicated to developing pharmacological and behavioral treatments for mental disorders over the years, a significant setback is symptom-based diagnoses that assume discrete disease categories, which may contribute to low remission rates and inadequate treatment responses as it does not fully account for the heterogeneity and complexity of mental disorders \citep{rush2006report}.
In response to the limitations inherent in traditional symptom-based diagnoses and to address significant between-patient heterogeneity, the National Institute of Mental Health (NIMH) advocates for a paradigm shift through the Research Domain Criteria (RDoC) initiative \citep{insel2010research}. This shift involves redefining mental disorders by identifying latent constructs derived from biological and behavioral measures across diverse domains of functioning (e.g., positive/negative affect, decision-making, social processing) at different levels (e.g., cells, genes,  and brain circuits), instead of solely relying on the clinical symptoms. 
Behavioral data collected from tasks such as the probabilistic reward task \citep[PRT;][]{pizzagalli2005toward} and the emotion conflict task \citep{etkin2006resolving} is a highly valued resource within the framework of RDoC. 
Utilizing carefully designed behavioral tasks with appropriate data- or theory-driven computational methods \citep{huys2016computational}, researchers can extract valuable behavioral markers that serve as indicators of mental health conditions.
Moreover, there is a critical need to examine the associations between the brain and behavior, as demonstrated by \cite{fonzo2019brain}. 
This exploration facilitates understanding the underlying brain mechanisms during behavioral tasks, allowing for the identification of brain abnormalities related to mental disorders, and ultimately contributes to refining diagnostic frameworks and developing more effective therapeutic interventions.

Reward learning and decision-making are pivotal in the etiology of mental function \citep{kendler1999causal}. Reward processing refers to how individuals perceive, value, and pursue rewards (e.g., money, social approval). MDD patients may show abnormalities in reward processing, such as reduced motivation, anhedonia (loss of pleasure), or reduced sensitivity to positive rewards. To measure reward processing, patients perform computer-based behavioral tasks that involve making choices or responding to stimulants that are associated with different outcomes, such as gains or losses in the laboratory. We focus on characterizing human decision-making behaviors through reward tasks (e.g., PRT), wherein individuals make decisions by interacting with a reward-based system and continually adapting their decision-making mechanisms based on obtained rewards.

Biological evidence suggests that fluctuations in human dopaminergic neurons indicate errors in predicting future rewarding events \citep{schultz1997neural}. This supports the use of simple prediction error learning (comparing real and expected reward) to update the value functions of different choices \citep{rescorla1972theory, huys2011disentangling}, laying the foundation for modeling behavior by Reinforcement Learning \citep[RL;][]{sutton2018reinforcement}.
To describe reward learning behaviors emerging from a reward learning task, \cite{huys2013mapping} introduced a prediction error RL approach to characterize decision making based on key behavioral phenotypes, including reward sensitivity and learning rate. In a PRT study, they identified reward sensitivity (instead of learning rate) as an important behavioral marker contributing to the abnormality in reward learning for MDD, with a lower reward sensitivity in MDD patients.
\cite{guo2023semiparametric} extends the approach of \cite{huys2013mapping} to a more general semiparametric RL framework by replacing the scalar value of reward sensitivity with a nonlinear and nondecreasing reward sensitivity function.
They revealed that the reward sensitivity for the MDD patients is not uniformly lower than the healthy subjects, but only when subjects receive sufficient rewards. 
Furthermore, \cite{guo2023semiparametric} noted 
a floor and ceiling effect in the relationship between choices and expected rewards, attributed to the non-linearity of the reward sensitivity function, indicating that the decision-making process may not adhere to a single RL model.
Several other studies also support that subjects employ multiple strategies in decision-making, as highlighted by \cite{worthy2013heterogeneity, iigaya2018effect, ashwood2022mice}.
In particular, \cite{ashwood2022mice} demonstrates that the perceptual decision-making process switches between multiple interleaved strategies. They suggest that the decision process involves a mixture of `engaged' and `lapse' strategies. During the `engaged' strategy, a subject makes choices following a generalized linear model with softmax. Conversely, during the `lapse' strategy, the subject ignores the stimulus and makes choices based on a fixed probability. The dynamics of strategy-switching are modeled by a hidden Markov model (HMM).
Alternatively, \cite{chen2021sex} considers `engaged' as exploitation and `lapse' as exploration, aligning with the exploitation-exploration trade-off in RL \citep{sutton2018reinforcement}. During the exploration stages, each choice is assigned equal probabilities.

In this work, we propose an RL-HMM framework to characterize reward-based decision-making with alternating strategies between two distinct decision-making processes: in the `engaged' state, subjects make decisions based on the RL model, while in the `lapse' state, subjects make decisions through random choices with equal probabilities.
Our approach generalizes the RL model introduced in \cite{huys2013mapping} and \cite{guo2023semiparametric} from finite discrete RL states to infinite but bounded state space.
We employ HMM with time-varying transition probabilities to model the decision-making strategy switching, extending the HMM framework in \cite{ashwood2022mice}, where the transition probabilities are assumed to be time-invariant. In our approach, the time-varying transition probabilities are modeled nonparametrically  with fused lasso regularization \citep{tibshirani2005sparsity} or trend filtering \citep{tibshirani2014adaptive}.
The proposed approach is applied to the Probabilistic Reward Task \citep{pizzagalli2005toward} in the EMBARC study \citep{trivedi2016establishing}, a large-scale randomized clinical trial recruiting participants diagnosed with MDD and healthy controls (details in Section \ref{sec:prt}).
{\color{black} 
To the best of our knowledge, this study is the first to consider the decision-making dynamics between RL and other decision-making strategies in probabilistic reward tasks. Compared to \cite{huys2013mapping} and \cite{guo2023semiparametric}, this work allows us to capture additional important features regarding the engagement patterns of both the MDD group and each individual subject in the task.
We demonstrate the importance of parameters obtained from our model, such as the group engagement rates and individual engagement scores (defined in Section \ref{sec:RL-HMM}). The group engagement rates reveal different decision-making strategy patterns between MDD patients and healthy subjects. The individual engagement scores serve as crucial behavioral markers that may be associated with specific brain regions or networks.
Understanding engagement patterns provides valuable insights into how patients interact with and adhere to treatment or intervention protocols in clinical trials. It can help healthcare providers and researchers optimize strategies to enhance patient involvement and ultimately improve treatment outcomes.}
{\color{black} 
Our proposed work aligns with the framework of imitation learning \citep{ross2010efficient} or behavioral cloning \citep{torabi2018behavioral}, wherein a policy is learned by mimicking the behavior demonstrated by an expert from their observed decision-making process. One difference is that we do not assume participants follow an optimal policy that maximizes their rewards. A related line of research is inverse reinforcement learning \citep{abbeel2004apprenticeship}.
}

The remainder of the article is organized as follows. Section \ref{sec:meth} introduces a general RL model with an approximate solution, outlines our RL-HMM method, and provides details on parameter estimation, inference, and model selection. In Section \ref{sec:simulation}, we present results from our simulation studies, while Section \ref{sec:realdata} focuses on the analysis of PRT data from the EMBARC study. In Section \ref{sec:discussion}, we summarize our approach, highlight key findings, and discuss potential extensions for future research.






%
%
%
%
%
%
%
\vspace{-3.5em}

\section{Methods}
\vspace{-0.75em}
\label{sec:meth}
Consider a group of $n$ subjects. The dataset comprises decision-making processes of each subject, (i.e.,
$
\dots, S_{it}, \ A_{it}, \ R_{it}, \ S_{it+1}, \ A_{it+1}, \ R_{it+1},\dots
$), 
where $S_{it}$, $A_{it}$, and $R_{it}$ represent the random variables of state, action, and reward for the $i$-th subject at time $t$, respectively. Define $\mathcal{H}_{it} = \{S_{ij}, A_{ij}, R_{ij}\}_{j=1}^{t-1}$ as the observed history for the $i$-th subject up to time $t$.
{\color{black} Throughout this paper, we assume that the state $S_{it}$ is generated from an arbitrary non-trivial distribution of the transitions, denoted as $P(S_{it} \mid \mathcal{H}_{it})$, with these states confined within a bounded state space denoted as $\mathcal{S}$. One special case is the contextual bandit problem, where states are generated independently, i.e.,  $P(S_{it} \mid \mathcal{H}_{it}) = P(S_{it})$.}
We consider a discrete action space $\mathcal{A}$ with $D$ available options (i.e. $\mathcal{A} = \{0, 1, \dots, D-1\}$). 
The decision-making models are introduced in detail in Sections \ref{sec:RL} to \ref{sec:RL-HMM}.
{\color{black} 
The reward $R_{it}$ is generated based on a function involving $S_{it}$, $A_{it}$, and $t$ (i.e., the reward-generating distribution is $P_t(R_{it} \mid S_{it}, A_{it})$), with the reward space $\mathcal{R} \subset [0, R_{\max}]$ being either discrete or continuous.
The mechanisms that generate rewards can be either stochastic or deterministic.
In this paper, our aim is to model human reward-based decision-making behavior in behavioral experiments where immediate rewards are provided at the end of each trial. Therefore, we do not consider downstream effects of rewards.}

\vspace{-2em}
\subsection{Decision-making with reinforcement learning} 
\vspace{-0.75em}
\label{sec:RL}


RL is a computational framework that can be used to model subjects' responses to rewards in behavioral tasks.
In this framework, subjects learn the underlying reward generating mechanism and subsequently make decisions based on their acquired knowledge. 

Define $Q_{i}^{t}(a, s)$ as the expected reward for taking action $a$ at state $s$ for the $i$th participant at the $t$th trial. 
Assume that the expected reward can be approximated by $\widetilde{Q}_i^t(a, s)=\bphi(a, s)^{\top} \bdel_i^t$, where $\bphi(a, s)$ represents the basis vector associated with the state $s \in \mathcal{S}$ for action $a \in \mathcal{A}$,  and $\bdel_i^t$ is the corresponding coefficient vector.
After observing the reward $R_{it} = r_{it}$ at trial $t$, the participant updates the expected reward $\widetilde{Q}_i^t(a, s)$ by adjusting it in the direction of $r_{it} - \widetilde{Q}_i^t(a, s)$. The corresponding update of the coefficients follows:
\vspace{-0.75em}
\begin{align}
    \bdel_i^{t+1} = \bdel_i^t + \beta \bphi(a_{it}, s_{it})\left[ r_{it}-\bphi(a_{it}, s_{it})^{\top} \bdel_i^t\right],
    \label{equ:update_delta}
\end{align}
where $\beta$ is the learning rate.
Note that \eqref{equ:update_delta} is also known as the Rescorla-Wagner equation \citep{rescorla1972theory}. It can be viewed as a one-step stochastic gradient descent, where we minimize the risk associated with the following mean squared error:
\vspace{-0.75em}
\begin{align}
\mathbb{E}_{R} \left[ \left( R_{it} - \bphi(a_{it}, s_{it})^{\top} \bdel_i^t \right)^2 \ \bigl\vert \ A_{it} = a_{it}, S_{it} = s_{it} \right]. \label{equ:MSE} 
\end{align}
Left multiplying $\phi^{\top}(a_{it},s_{it})$ to both sides of \eqref{equ:update_delta}, we obtain
\vspace{-0.75em}
\begin{align}
    \wt{Q}_i^{t+1}(a_{it},s_{it})=\beta_{it} r_{it} + (1-\beta_{it}) \wt{Q}_i^t(a_{it},s_{it}),
    \label{equ:update_Q}
\end{align}
where $\beta_{it} = \beta \bphi(a_{it}, s_{it})^{\top} \bphi(a_{it}, s_{it})$. To ensure that the expected reward at $t + 1$ is the weighted sum of the obtained reward and the expected reward at $t$, the learning rate should satisfy $0 < \beta_{it} < 1$.
As a note, if $s$ is discrete and we choose $\phi(a,s)$ as indicator functions on all discrete values in $\mathcal{A} \times \mathcal{S}$, then $\bphi(a_{it}, s_{it})^{\top} \bphi(a_{it}, s_{it})\equiv 1$ and $\beta_{it} \equiv \beta$ in \eqref{equ:update_Q}.
Alternatively, if $s$ is continuous,  we choose $\phi(a,s) = (I(a=0), I(a=1))^{\top} \otimes \psi(s)$, where $\psi(s)$ can be a linear basis or B-splines.
We further assume that $\boldsymbol{\alpha} := \boldsymbol{\delta}^{1} = \boldsymbol{\delta}_i^{1}$.
We assume that the action $A_{it}$ follows the model
\vspace{-0.5em}
\begin{align}
    P\left(A_{it}=a \mid S_{it}, \mathcal{H}_{it} \right) = \frac{\exp\{ \nu_a + \rho Q_i^t(a, S_{it}) \}}{\sum_{a'=0}^{D-1} \exp\{ \nu_{a'} + \rho Q_i^t(a', S_{it}) \} } \label{equ:prob_RL},
\end{align}
where $\rho > 0$ is the reward sensitivity. A higher reward sensitivity indicates that subjects' actions will depend more on the expected rewards they learned. Here, $\nu_a$ $(a=0, \dots, D-1)$ are time-invariant intercepts. To ensure identifiability, we let $\nu_0 \equiv 0$, and $\nu_1 = \dots = \nu_{D-1} = 0$ implies that the subject is making decisions entirely based on the expected reward.
{\color{black} 
The softmax decision-making model \eqref{equ:prob_RL} is widely used in policy gradient methods \citep{sutton2018reinforcement} in RL because it is differentiable with respect to its parameters and its stochastic nature ensures exploration.}

\vspace{-3.5em}
\subsection{Decision-making with state switching strategies}
\label{sec:RL-HMM}

Several studies provide evidence that subjects employ multiple strategies for decision-making \citep{worthy2013heterogeneity, iigaya2018effect, ashwood2022mice}. 
In this work, we consider that subjects exhibit two distinct decision-making strategies: the `engaged' state (represented as $U_{it} = 1$), in which subjects make decisions based on the RL model \citep{huys2013mapping,guo2023semiparametric}, and the `lapse' state (represented as $U_{it} = 0$), where subjects make decisions through random choices, with an equal probability of $1/D$ for selecting one of the $D$ available actions \citep{chen2021sex}.
Here, $U_{it} \in \{0, 1\}$ represents a latent variable that indicates the decision-making strategy used by the subject $i$ during trial $t$.
{\color{black} We assume that in the behavioral decision-making framework, how subjects make decisions is independent of how they learn from observed (action, reward) pairs.
Even in the `lapse' phase, subjects still internally process reward updates by referencing the expected reward output, akin to having a computational mechanism in their brain. 
Engagement, however, depends on various factors such as mood, emotions, and potentially depression, which can vary significantly. For instance, a significant issue in mobile health is that some patients disengage from their studies due to fatigue induced by prompts.}
We introduce a novel RL-HMM framework for the analysis of perceptual decision-making behavior, employing a combination of hidden Markov model (HMM) and reinforcement learning. The proposed RL-HMM model is: 
\vspace{-1em}
\begin{align}
    P\left(A_{it}=a \mid U_{it}, S_{it}, \mathcal{H}_{it} \right) &= I(U_{it}=1)\frac{\exp\{ \nu_a + \rho Q_i^t(a, S_{it}) \}}{\sum_{a'=0}^{D-1} \exp\{ \nu_{a'} + \rho Q_i^t(a', S_{it}) \} } + I(U_{it}=0) \frac{1}{D}, \\
    P\left(U_{i,t+1}=1 \mid U_{it}=j \right) &= \frac{\exp(\zeta_{jt})}{1 + \exp(\zeta_{jt}) } \quad j=0, 1.
\end{align}
 Here, $\zeta_{jt} = \zeta_{j}(t)$ represents a nonparametric function indicating that the transition probability may vary across trials. 
 The decision-making strategy at the initial trial is characterized by the probability distribution $P\left(U_{i1} = j\right) = \pi_j$, where $0 < \pi_0, \pi_1 < 1$, and $\pi_0 + \pi_1 = 1$.

{\color{black} Define $\mathbf{A}_i = \mathbf{A}_{i,[1:T]} = (A_{i1}, \dots, A_{iT})$ (a similar definition applies to $\mathbf{S}_i, \mathbf{R}_i$ and $\mathbf{U}_i$). We can predict the probability of $(U_{it}=1)$ by its posterior $\gamma_{i1t} = P\left(U_{it}=1 \mid \mathbf{A}_i\right)$; we refer to this quantity as the individual engagement probability at trial $t$. The trajectories can exhibit the learning pattern of an individual.
We define the group engagement rate at trial $t$ as the average of the individual engagement probabilities in the group, i.e., $n^{-1} \sum_{i=1}^n {\gamma}_{i1t}$.
We define the individual engagement score as $\mbox{ES}_i(\mathcal{T}) = \frac{1}{|\mathcal{T}|} \sum_{t\in \mathcal{T}} \log(\gamma_{i1t} / (1 - \gamma_{i1t} ))$. This score quantifies the average (logit transformation of) individual engagement probabilities over trials in the trial set $\mathcal{T}$.}

\vspace{-1.5em}
\subsection{Parameter estimation via EM algorithm}\label{sec:computation}
\vspace{-0.75em}

In the proposed model, let $\boldsymbol{\theta} = \left\{ \beta, \rho, \nu_1, \dots, \nu_{D-1}, \balpha, \pi_1, \boldsymbol{\zeta}_0, \boldsymbol{\zeta}_1 \right\}$ denote the collection of parameters of interest, where $\boldsymbol{\zeta}_j = \zeta_{j,[1:T-1]}$.
The joint likelihood of $(\mathbf{A}_{1:n}, \mathbf{U}_{1:n}, \mathbf{S}_{1:n}, \mathbf{R}_{1:n})$ for RL-HMM takes the following form
\vspace{-0.5em}
\begin{align*}
    \prod_{i=1}^n \prod_{t=1}^T P\left(U_{it} \mid U_{it-1}; \boldsymbol{\theta}  \right) P\left(S_{it} \mid \mathcal{H}_{it} \right) P\left(A_{it} \mid U_{it}, S_{it}, \mathcal{H}_{it}; \boldsymbol{\theta} \right) P_t\left(R_{it} \mid S_{it}, A_{it} \right) ,
\end{align*}
where $P\left(U_{i1} \mid U_{i0} \right) = P\left(U_{i1}\right)$.
Rather than focusing on the learning process of how rewards and states are generated, which are the goals of the participants, the primary objective of this paper is to model human reward-based decision-making behaviors in behavioral tasks. 
Consequently, the parameters in the distributions generating states and rewards (i.e., $P\left(S_{it} \mid \mathcal{H}_{it} \right)$ and $P_t\left(R_{it} \mid S_{it}, A_{it} \right)$) are distinct from the parameter $\boldsymbol{\theta}$, so that the log-likelihood function to factorize into $\mathcal{L}_{n}(\boldsymbol{\theta})$ and a term that does not depend on $\boldsymbol{\theta}$. $\mathcal{L}_{n}(\boldsymbol{\theta})$ can be expressed as follows:
\begin{align*}
    \mathcal{L}_n\left(\boldsymbol{\theta}\right) &= \sum_{i=1}^n \bigg( \log P(U_{i1}) + \sum_{t=1}^{T-1}  \log P\left(U_{i,t+1} \mid U_{it} \right)  + \sum_{t=1}^T \log P\left(A_{it} \mid U_{it}, S_{it}, \mathcal{H}_{it} \right)  \bigg).
\end{align*}
To avoid overfitting, we regularize $\boldsymbol{\zeta}_j$ via trend filtering penalty \citep{tibshirani2014adaptive} with form:
$\lambda_0 \left\| \mathbf{D}^{(r+1)} \boldsymbol{\zeta}_0 \right\|_1 + \lambda_1 \left\| \mathbf{D}^{(r+1)} \boldsymbol{\zeta}_1 \right\|_1$, where $\mathbf{D}^{(r+1)} \in \mathbb{R}^{(T-r-2) \times (T-1)}$ is the discrete difference operator of order $r$. The definition of $\mathbf{D}^{(r+1)}$ can be found in the Supplementary Materials Section S.1.    
When $r=0$, the zero-th order trend filtering problem reduces to one-dimensional fused Lasso \citep{tibshirani2005sparsity}. When $r=0$ and $\lambda_0, \lambda_1 \rightarrow \infty$, the estimated HMM will have fixed transition probabilities for all trials. We estimate $\btheta$ by the EM algorithm.

\noindent \textbf{E-step}:
The E-step involves computing the posterior distribution of $U_{it}$, which is challenging due to the HMM structure. The forward-backward algorithm \citep{baum1970maximization, ashwood2022mice} addresses the challenge by employing recursion and memoization, with both the forward pass and backward pass of the algorithm requiring just a single traversal through all trials within a session.  Specifically, we first compute the expectation of $\mathcal{L}_n\left(\boldsymbol{\theta} \right)$ in terms of $P(\mathbf{U}_{1:n} \mid \mathbf{A}_{1:n}, \boldsymbol{\theta}^{\mbox{old}})$:
\begin{align}
    \mathrm{E}_{U}\left( \mathcal{L}_n\left(\boldsymbol{\theta} \right) \mid \mathbf{A}_{1:n}, \boldsymbol{\theta}^{\mbox{old}} \right) = & \sum_{i=1}^n \Bigg\{ \sum_{j=0}^1 \gamma_{ij1}(\btheta^{\mbox{old}}) \cdot \log \pi_{j}  
     + \sum_{t=1}^{T-1} \sum_{j=0}^1 \sum_{k=0}^1 \xi_{ijkt}(\btheta^{\mbox{old}}) \cdot \log C_{jk}^t(\btheta) \nonumber \\
    & \quad\quad + 
    \sum_{t=1}^T \sum_{j=0}^1 \gamma_{ijt}(\btheta^{\mbox{old}}) \cdot \log \eta_{ijt}(\btheta)   \Bigg\}. \label{equ:exp_joint_likelihood}
\end{align}
Here, $\gamma_{ijt} = \gamma_{ijt}(\btheta)$, $\xi_{ijkt} = \xi_{ijkt}(\btheta)$, $\eta_{ijt} = \eta_{ijt}(\btheta)$, and $C_{jk}^t = C_{jk}^t(\btheta)$ are defined by
\vspace{-0.75em}
\begin{align*}
    &\gamma_{ijt}(\btheta) = P\left(U_{it}=j \mid \mathbf{A}_i, \boldsymbol{\theta} \right), 
    \quad\quad\quad
    \xi_{ijkt}(\btheta) = P\left(U_{i,t+1}=k, U_{it}=j \mid \mathbf{A}_i, \boldsymbol{\theta} \right), \\
    &\eta_{ijt}(\btheta) = P\left(A_{it} \mid U_{it}=j, S_{it}, \mathcal{H}_{it}, \btheta \right),
    \quad
    C^t_{jk}(\btheta) = P\left(U_{i,t+1}=k \mid U_{it}=j, \btheta \right).
\end{align*}
Define $a_{ijt} = P\left(U_{it}=j, \mathbf{A}_{i,[1:t]} \right)$,
we can compute the value of $a_{ijt}$ through forward passing with $a_{ij1} = \pi_{j} \eta_{ij1}$ for $t=1$ and $a_{ijt} = \sum_{k=0}^1 a_{ik,t-1} \eta_{ijt} C_{kj}^{t-1}$ for $t =2, \dots, T$.
Define $b_{ijt} = P\left(\mathbf{A}_{i,[t+1:T]} \mid U_{it}=j, \mathbf{A}_{i,[1:t]} \right)$, we can compute the value of $b_{ijt}$ through backward passing with $b_{ijT} = 1$ for $t=T$ and $b_{ijt} = \sum_{k=0}^1 b_{ik,t+1} \eta_{ik,t+1} C_{jk}^t$ for $t=T-1, \dots, 1$.
The derivation of $a_{ijt}$ and $b_{ijt}$ uses the Markov property of $U_{it}$. 
Then we compute the posterior distribution of $U_{it}$ by
\vspace{-0.5em}
\begin{align*}
    \gamma_{ijt}  = \frac{a_{ijt} b_{ijt}}{\sum_{k=0}^1 a_{ikT}} \quad\quad \mbox{and} \quad\quad 
    \xi_{ijkt} =  \frac{ a_{ijt} b_{ik,t+1} \eta_{ik,t+1} C_{jk}^t }{ \sum_{k=0}^1 a_{ikT} }. 
\end{align*}
The derivation details of $a_{ijt}$, $b_{ijt}$, $\gamma_{ijt}$, and $\xi_{ijkt}$ are provided in the Supplementary Materials.

\noindent \textbf{M-step}: write the objective function as follow:
\vspace{-0.75em}
\begin{align}
    \mathcal{J}_{\boldsymbol{\lambda}}\left(\boldsymbol{\theta} \right) &=
    -\mathrm{E}_{U}\left( \mathcal{L}_n\left(\boldsymbol{\theta} \right) \mid \mathbf{A}_{1:n}, \boldsymbol{\theta}^{\mbox{old}} \right) + \sum_{j=0}^1 \lambda_j \left\| \mathbf{D}^{(r+1)} \boldsymbol{\zeta}_j \right\|_1,
    \label{equ:object_func}
\end{align}
where $\boldsymbol{\lambda} = (\lambda_0, \lambda_1)$.
From \eqref{equ:exp_joint_likelihood}, we find that we can partition $\mathcal{J}_{\boldsymbol{\lambda}}\left(\boldsymbol{\theta} \right)$ into four components that only contain $\pi_1$, $\boldsymbol{\zeta}_0$, $\boldsymbol{\zeta}_1$, $(\beta, \rho, \nu_1, \dots, \nu_{D-1}, \boldsymbol{\alpha})$, respectively. Each part is minimized separately. 
We apply the generalized EM algorithm \citep{dempster1977maximum}, where we replace the minimization step with a one-step update of Newton's method. At the $(l+1)$-th iteration, we update the parameter estimates as follows:

\noindent \textbf{Update $\pi_1$}:
${\pi_1}^{(l+1)} = n^{-1} \sum_{i=1}^n \gamma_{i11}(\btheta^{(l)})$.

\noindent \textbf{Update $\boldsymbol{\zeta}_j$ $(j=0,1)$}: Define $\xi_{ijkt}^{(l)} = \xi_{ijkt}(\btheta^{(l)})$
write \eqref{equ:object_func} in terms of $\boldsymbol{\zeta}_j$
\begin{align*}
    \mathcal{J}_{\lambda_j}\left( \boldsymbol{\zeta}_{j} \right) &= \sum_{i=1}^n \sum_{t=1}^{T-1} \bigg\{ -\xi_{ij1t}^{(l)} \zeta_{jt} + \left( \xi_{ij0t}^{(l)} + \xi_{ij1t}^{(l)} \right) \log\left( 1 + \exp\left( \zeta_{jt} \right) \right)  \bigg\} + \lambda_j \left\| \mathbf{D}^{(r+1)} \boldsymbol{\zeta}_j \right\|_1.
    \end{align*}
When $\lambda_j=0$, the gradient $\mathbf{g}_0(\boldsymbol{\zeta}_j) = (g_0(\zeta_{j1}), \dots,  g_0(\zeta_{j,T-1}))^{\top}$ and Hessian matrix $\mathbf{H}_0(\boldsymbol{\zeta}_j) = \mbox{diag}(H_0(\zeta_{j1}), \dots,  H_0(\zeta_{j,T-1}))$ of $\mathcal{J}_{0}$ in terms of $\boldsymbol{\zeta}_{j}$ can be represented by
\begin{align*}
& g_0(\zeta_{jt}) = \sum_{i=1}^n \Big\{ -\xi_{ij1t}^{(l)} + \left(\xi_{ij0t}^{(l)} + \xi_{ij1t}^{(l)} \right) C_{j1}^t \Big\}, \quad
H_0(\zeta_{jt}) = \sum_{i=1}^n  \left(\xi_{ij0t}^{(l)} + \xi_{ij1t}^{(l)} \right) C_{j1}^t (1-C_{j1}^t).
\end{align*}
We update ${\bzeta}_j^{(l+1)}$ using the least-square generalized lasso problem \citep{tibshirani2011solution}:
\begin{align}
    {\bzeta}_j^{(l+1)} = \mathbf{H}_0^{-1/2}(\boldsymbol{\zeta}_{j}^{(l)} ) \widetilde{\bzeta}_j^{(l+1)}, \quad
    \widetilde{\bzeta}_j^{(l+1)} = \mbox{argmin}_{ \widetilde{\bzeta}_j } \frac{1}{2} \| \mathbf{y}_j^{(l)} - \widetilde{\bzeta}_j \|_2^2 + \lambda_j \| \widetilde{\mathbf{D}}_{r+1}^{(l)} \widetilde{\bzeta}_j \|_1, \label{equ:gen_lasso1}
\end{align}
where $\mathbf{y}_j^{(l)} = \mathbf{H}_0^{1/2}(\boldsymbol{\zeta}_{j}^{(l)} ) \boldsymbol{\zeta}_{j}^{(l)} - \mathbf{H}_0^{-1/2}(\boldsymbol{\zeta}_{j}^{(l)} ) \mathbf{g}_0(\boldsymbol{\zeta}_{j}^{(l)})$, $\widetilde{\mathbf{D}}_{r+1}^{(l)} = \mathbf{D}^{(r+1)} \mathbf{H}_0^{-1/2}(\boldsymbol{\zeta}_{j}^{(l)} )$.
The derivation of \eqref{equ:gen_lasso1} can be found in the Supplementary Materials Section S.1. The least-square generalized lasso \eqref{equ:gen_lasso1} can be efficiently solved using the R package \textit{genlasso} \citep{arnold2016efficient}.

\noindent \textbf{Update $(\beta, \rho, \nu_1, \dots, \nu_{D-1}, \boldsymbol{\alpha})$}:
Write \eqref{equ:object_func} 
\begin{align*}
    \mathcal{J}_{\boldsymbol{\lambda}}\left(\beta, \rho, \nu_1, \dots, \nu_{D-1},  \boldsymbol{\alpha} \right) &= \sum_{i=1}^n \sum_{t=1}^{T} \gamma_{i1t}(\btheta^{(l)}) \bigg\{ - \sum_{a=0}^{D-1} I(A_{it}=a) \left(\nu_a + \rho Q_i^t(a, S_{it}) \right) \\
    & \quad \quad \quad \quad + \log \Big( \sum_{a=0}^{D-1} \exp\left\{ \nu_a + \rho Q_i^t(a, S_{it}) \right\} \Big) \bigg\}.
\end{align*}
The L-BFGS-B algorithm \citep{byrd1995limited} is implemented to find the one-step update of $\beta^{(l+1)}$, $\rho^{(l+1)}$, $\nu_a^{(l+1)}$, and $\balpha^{(l+1)}$ where we specify the constraints $\rho > 0$ and $0< \beta < 1$.
A numerical approximation is employed to obtain the gradient of $\mathcal{J}_{\boldsymbol{\lambda}}$ in each direction.

\subsection{Parameter inference and model evaluation}

In addition to point estimation, assessing the variability of parameter estimates is essential for statistical inference. 
{\color{black} We employ nonparametric bootstrapping inference, treating the reward learning process for each subject $\mathcal{H}_{i,T+1} = (\mathbf{S}_{i}, \mathbf{A}_{i}, \mathbf{R}_{i})$ as a unified entity, which is then resampled with replacement. Hence, the bootstrap method is deemed valid as $\mathcal{H}_{i,T+1}$ $(i=1, \dots, n)$ are assumed to be independent and identically distributed (i.i.d.) when $n$ is large.}
We construct $95\%$ bootstrap confidence intervals under the normal
distribution (i.e., $ \mbox{MLE} \pm 1.96 \times \mbox{bootstrap se.}$) in simulation studies and data application.
Since the distribution of $\pi_1$ is left-skewed when the true $\pi_1 > 0.5$ and right-skewed when the true $\pi_1 < 0.5$, we apply a logit transformation to $\pi_1$ before using a normal approximation to construct confidence intervals.

One way to evaluate the performance of a model is by utilizing the $K$-fold cross-validation and computing the out-of-sample observed joint log-likelihood.
For subject $i$, 
recall the observed likelihood takes form: $P( \mathbf{A}_{i,[1:T]}; \btheta )=\sum_{j=0}^1 a_{ijT}(\btheta)$.
Thus, the $K$-fold cross-validation score is
\vspace{-0.5em}
\begin{align}
    \frac{1}{nT}\sum_{k=1}^K \sum_{i \in \mathcal{I}_k} \log \sum_{j=0}^1 a_{ijT}(\widehat{\btheta}^{(-k)}), \label{equ:cv_score}
\end{align}
where $\mathcal{I}_k$ represents the $k$-th leave-out set, and $\widehat{\btheta}^{(-k)}$ is the estimate of $\btheta$ obtained using the remaining $K-1$ training sets. We select the penalty parameters $\lambda_0$ and $\lambda_1$ by maximizing \eqref{equ:cv_score}.

\vspace{-1em}
\section{Simulation studies}\label{sec:simulation}
\vspace{-0.75em}

We conducted extensive simulation studies to assess the finite-sample performance of the proposed method. 
For the RL model, we considered a continuous state space $\mathcal{S} = [0, 1]$ and a two-arm action space $\mathcal{A} = \{0, 1\}$. 
We considered the reward-generating distribution as $p(R_{it} \mid A_{it} = a, S_{it} = s) \sim \mbox{Bernoulli} \big(p_a\{as+(1-a)(1-s)\} \big)$. We let $p_0 = 0.5$, $p_1 = 1$. The rewards at different time points were assigned independently. The states were independently generated by uniform distribution $\mbox{U}(0, 1)$. We set learning rate $\beta = 0.05$, reward sensitivity $\rho = 4$, and intercept $\nu_1 = 0$. We choose a linear basis $\phi^{\top}(a,s) = (I(a=0), I(a=1)) \otimes (1, s)$ and assume that the initial coefficient has the form $\balpha = (\alpha/\rho) \cdot (1, -1, 0, 1)$, where we set $\alpha = 2$. 
We considered two HMM cases: switching between engaged and lapse (Case I) and always engaged (Case II).
For Case I, suppose $\pi_1 = 0.8$, the transition probabilities are piecewise constant functions with $\zeta_{0t} = -1.5 \cdot (1 + I(T/2 \le t < T))$ and $\zeta_{1t} = 2 \cdot (1 + I(T/2 \le t < T))$. For Case II, there is no decision-making strategy switching.

We generated 200 replicates with sample sizes $n \in \{100, 200\}$ and the number of trials $T \in \{100, 200\}$. We estimated the parameters of interest, $\btheta = \left\{ \beta, \rho, \nu_1, \alpha, \pi_1, \boldsymbol{\zeta}_{0}, \boldsymbol{\zeta}_{1} \right\}$, using the proposed EM algorithm with a fused Lasso penalty.
The regularization parameters $(\lambda_0, \lambda_1)$ were selected by maximizing the $5$-fold cross-validation scores in \eqref{equ:cv_score}, where the values of the regularization parameters were chosen from $+\infty$ to $0$, at which the solution path of generalized Lasso changes slope \citep{arnold2016efficient}.

For Case I, we used $50$ bootstrap samples to construct $95\%$ bootstrap confidence intervals for the parameters under the normal distribution. The evaluation of $\boldsymbol{\zeta}_{0}$ and $\boldsymbol{\zeta}_{1}$ was conducted at $t=T/4$ and $t=3T/4$, respectively.
The simulation results of the parameter estimate from $200$ replicates were given in Table \ref{table:1}. The estimate bias (Bias), standard deviation (SD), average bootstrap standard error of the $50$ bootstrap samples (SE), and coverage probability of the $95\%$ bootstrap confidence intervals (CP) were reported. Furthermore, we compared three approaches: (i) our proposed method (RL-HMM), (ii) the RL-HMM model with fixed transition probabilities over trials (RL-HMM-fixed), and (iii) the RL model without decision-making strategy switching (RL-only).
Table \ref{table:1} demonstrates that when subjects' decision-making involves a mixture of two decision-making strategies, the RL-only model tends to underestimate the  reward sensitivity ($\rho$) and the initial coefficient of Q values ($\alpha$) in comparison to the RL-HMM. Additionally, when the transition probabilities are time-varying, employing RL-HMM-fixed results in a larger absolute bias for most of the parameters.
Meanwhile, for our proposed RL-HMM, the bias of most parameters decrease when $n$ and $T$ increase.
Furthermore, the estimates of $\zeta_{0t}$ and $\zeta_{1t}$ are biased due to the presence of generalized lasso penalty terms, indirectly causing bias in $\rho$.
The coverage probabilities of all parameter estimates are close to the nominal level ($95\%$) for all three cases.
In the Supplementary Materials Section S.2, we show that 200 replicates and 50 bootstrap samples are sufficient to provide accurate point estimates and confidence intervals.

For Case II, the simulation results of the parameter estimates from $200$ replicates are presented in Table S1 in the Supplementary Materials. Note that cross-validation tends to favor time-invariant transition probabilities for the RL-HMM, making it equivalent to RL-HMM-fixed in Case II. Therefore, we only present RL-HMM in Table S1.
The observations indicate that when the underlying true mechanism is an RL model without decision-making strategy switching, both the RL-only model and RL-HMM exhibit a small bias for $\beta$ and $\nu_1$. However, the RL-HMM model shows a relatively larger bias for $\rho$ and $\alpha$, and this bias diminishes as $n$ and $T$ increase. Specifically, for $(T=200, n=200)$, the relative bias (i.e., $\text{bias} / \text{truth}$) for both $\rho$ and $\alpha$ is smaller than $2\%$.

\begin{figure}[!p]
    \centering
        \begin{tabular}{cc}
          \includegraphics[scale=0.21]{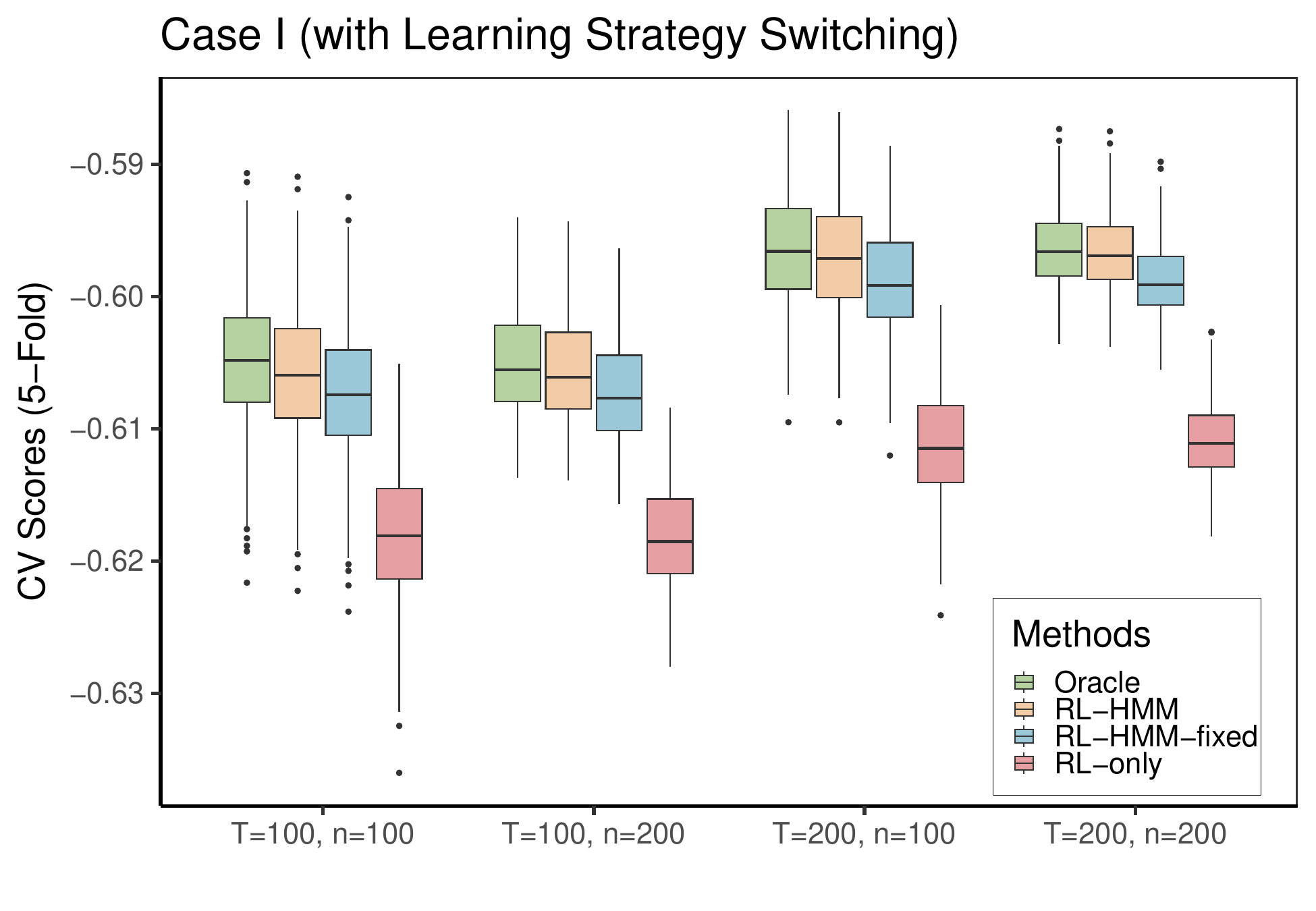} & 
          \includegraphics[scale=0.21]{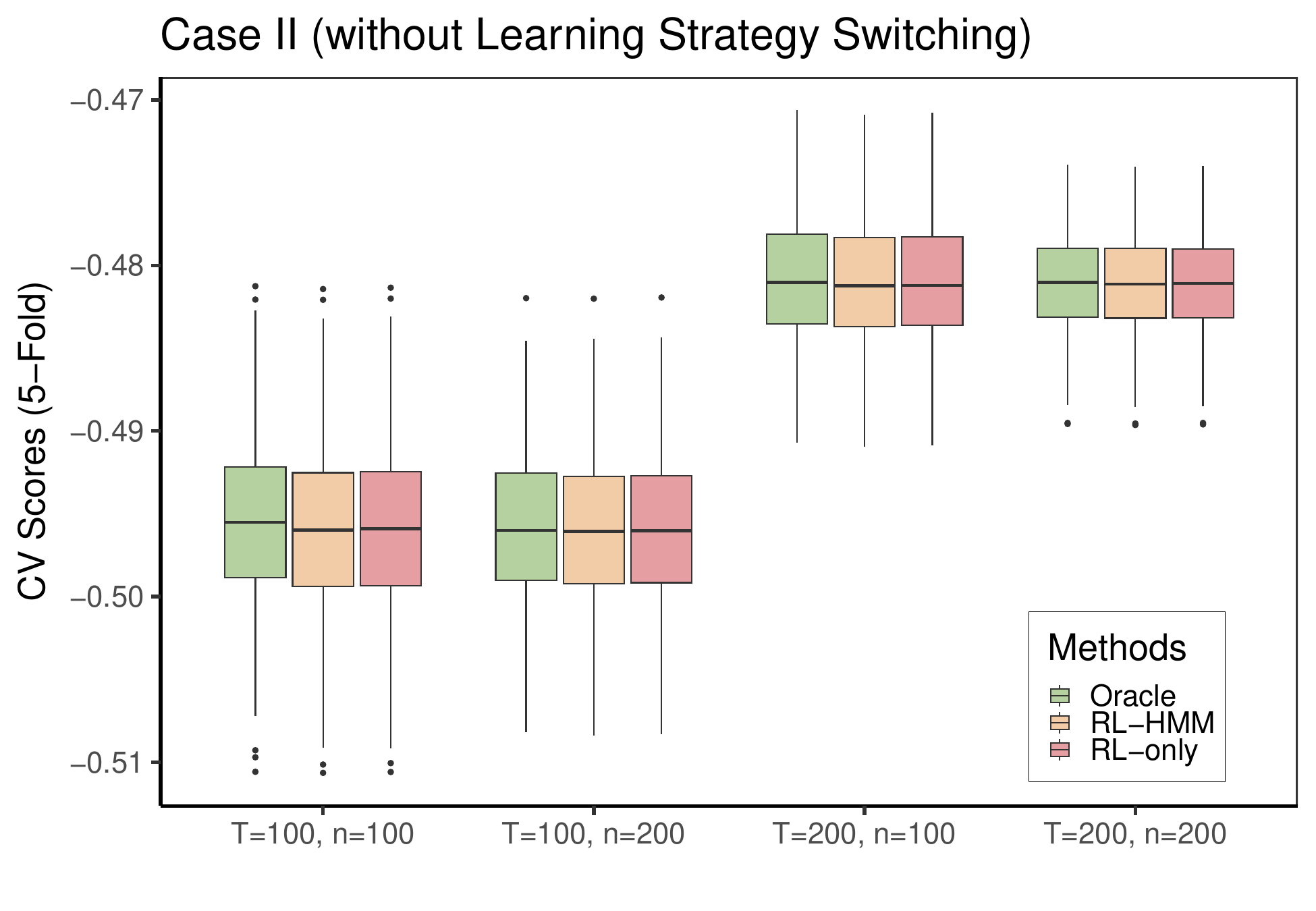} \\
          (a) Cross-validation scores & (b) Cross-validation scores   \vspace{2em} \\
         
          \includegraphics[scale=0.21]{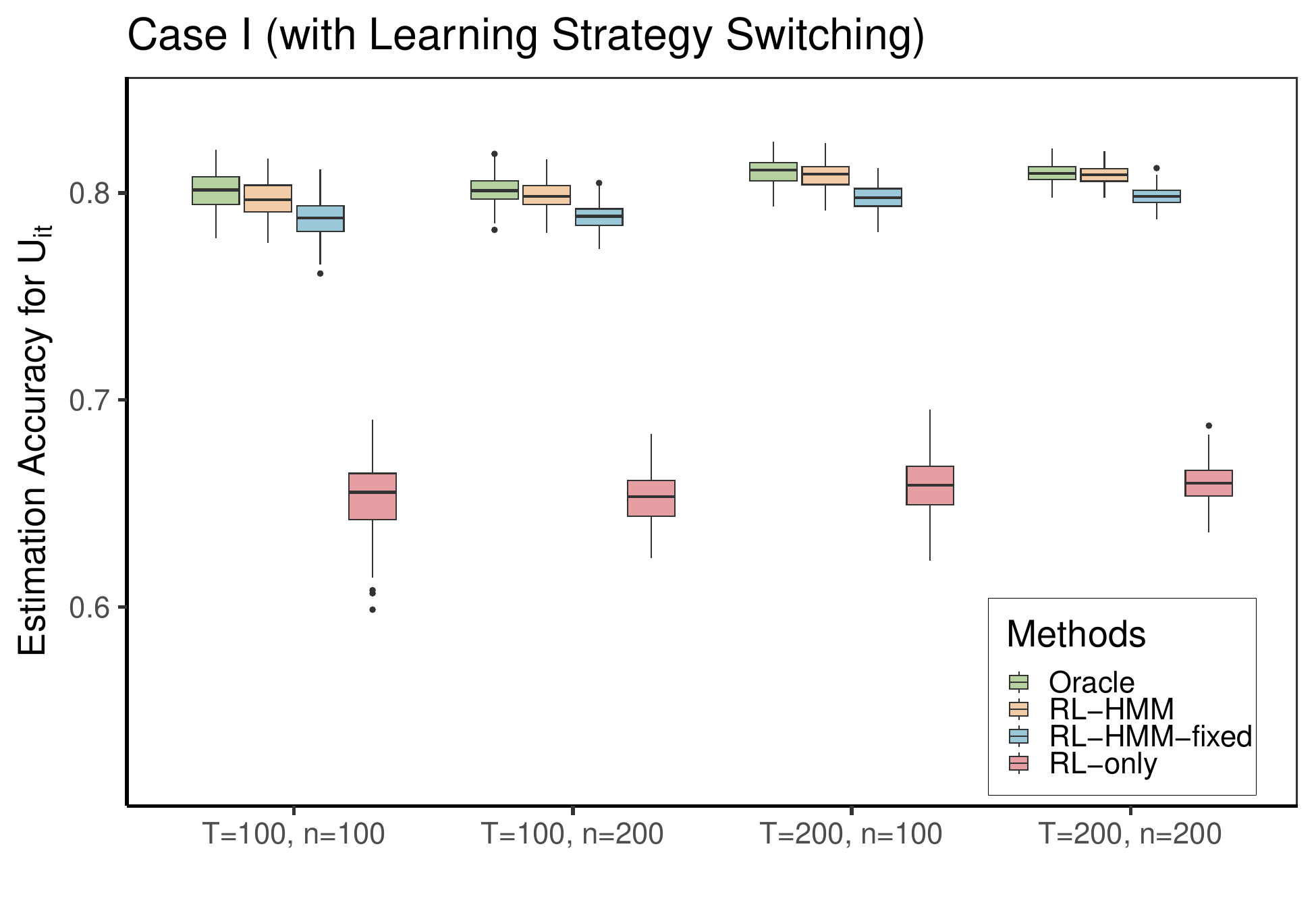} & 
          \includegraphics[scale=0.21]{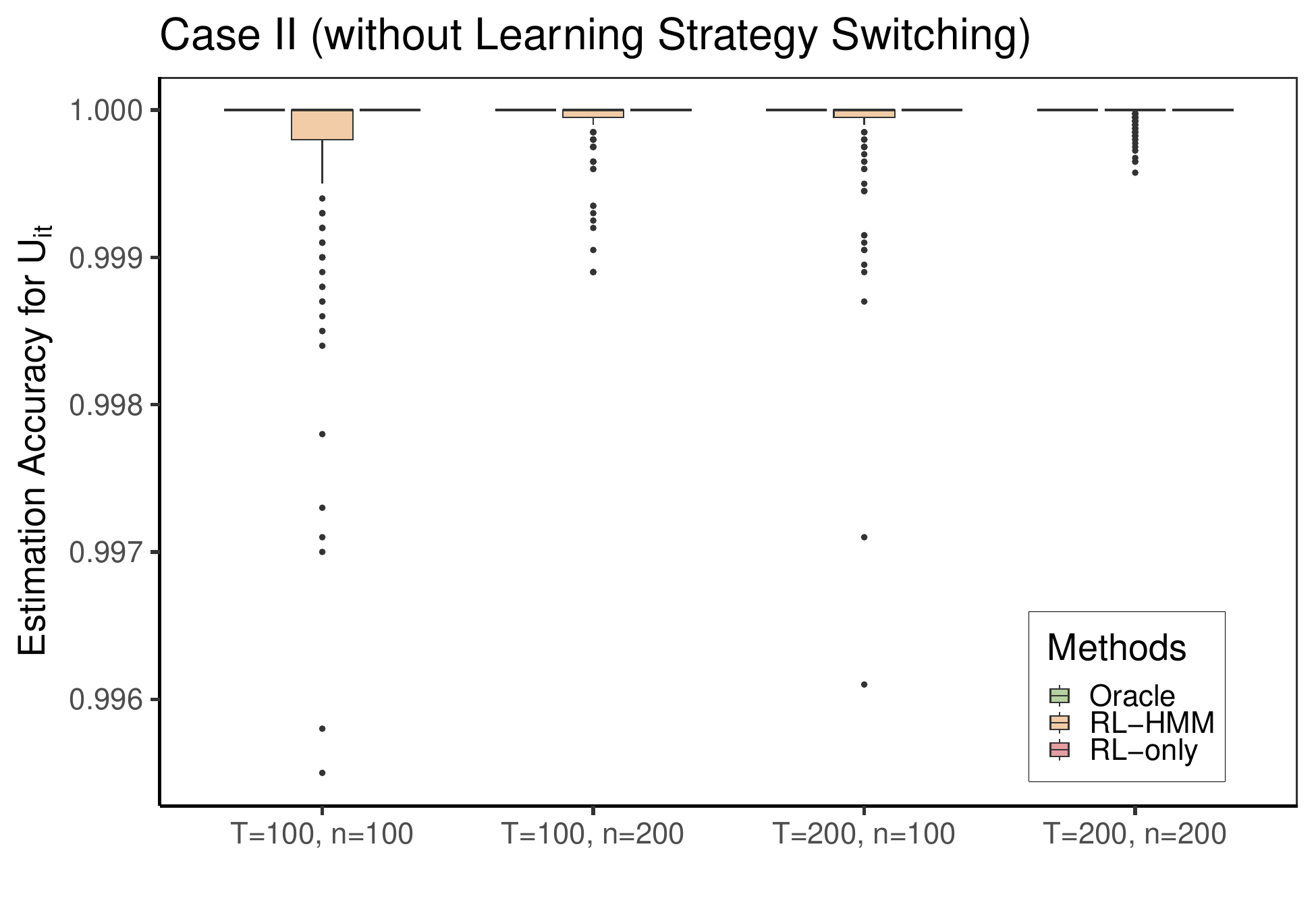} \\
          (c) Accuracy for estimating $U_{it}$ & (d) Accuracy for estimating $U_{it}$  \\
        \end{tabular}
    \caption{Boxplots show 5-fold cross-validation scores over the 200 replicates in Case I (a) and Case II (b), and boxplots show the estimation accuracy for the decision-making strategy over 200 replicates in Case I (c) and Case II (d). Models compared include the true model (Oracle), our proposed method (RL-HMM), RL-HMM model with time-invariant transition probabilities (RL-HMM-fixed), and the RL model without decision-making strategy switching (RL-only).
    The results for RL-HMM-fixed is not presented for Case II because cross-validation tends to select RL-HMM with time-invariant transition probabilities, making it equivalent to RL-HMM-fixed.}
    \label{img:simu}
\end{figure}

Finally, in both Case I and Case II, we presented the 5-fold cross-validation (CV) scores for the 200 replicates using boxplots in Figure \ref{img:simu} (a) and (b). The oracle (i.e., parameters are known) is compared with the RL-HMM, RL-HMM-fixed, and the RL model. In Figure \ref{img:simu} (a) for Case I, we observed that the RL-HMM demonstrates nearly the same performance as the oracle and outperforms RL-HMM-fixed, with RL showing the least favorable CV performance.
In Figure \ref{img:simu} (b) for Case II, all models exhibit similar CV performance.
On the other hand, recalling the estimated individual engagement probability for subject $i$ at trial $t$ as $\widehat{\gamma}_{i1t} = P(U_{it}=1 \mid \mathbf{A}_i, \widehat{\btheta} )$, we estimate the decision-making strategy at trial $t$ for subject $i$ by $\widehat{U}_{it} = I(\widehat{\gamma}_{i1t} \ge 0.5)$.
We present the averaged accuracy for estimating $U_{it}$ (i.e., $(nT)^{-1} \sum_{i=1}^n \sum_{t=1}^T I(\widehat{U}_{it} = U_{it})$) for the 200 replicates using boxplots in Figure \ref{img:simu} (c) and (d). In Figure \ref{img:simu} (c) for Case I, the RL-HMM outperforms RL-HMM-fixed, with RL showing the lowest accuracy because the RL model classifies every $U_{it}$ as `engaged'.
In Figure \ref{img:simu} (d) for Case II, RL-HMM estimates more than $99.9\%$ of $U_{it}$ correctly for most of the replicates, demonstrating the robustness of RL-HMM when the model is misspecified.
In the Supplementary Materials Section S.2, we employ additional simulations to demonstrate the robustness of our proposed method to the misspecification of the basis function $\bphi(a,s)$ and the choice of trend filtering order used to fit the nonparametric function $\zeta_{jt}$.

\vspace{-2.5em}
\section{Application to EMBARC Study}\label{sec:realdata}

\vspace{-0.8em}
\subsection{Probabilistic reward task and EMBARC study}\label{sec:prt}
In the EMBARC study \citep{trivedi2016establishing},
the PRT, as introduced by \cite{pizzagalli2005toward}, is a computer-based experiment used to assess subjects' ability to adapt their behavior in response to rewards. During each trial of the task, participants are presented with a cartoon face displaying one of the two stimuli (i.e., a face with a short mouth or with a long mouth). The objective for participants is to identify the type of stimulus in each trial.
Crucially, the PRT provides a higher frequency of rewarding participants for correctly identifying faces with a short mouth, designating it as the `rich reward', while providing less frequent rewards for correctly identifying faces with a long mouth, referred to as the `lean reward'.
Participants receive verbal instructions outlining the task's goal, which is to maximize the total rewards. It is essential that participants understand that not all correct responses will be rewarded. To optimize their rewards, participants should focus on providing the correct response, regardless of the face associated with the higher reward. However, 
the difference in mouth length between the short and long mouths is intentionally kept small.
The subtle difference in mouth length often leads to difficulties in accurately perceiving the stimuli presented, resulting in a tendency to prioritize states with higher rewards over those that are more accurate.

In the EMBARC study, each PRT session comprised 200 trials, divided into two blocks of 100 trials, with a 30-second break separating the blocks. In each block, 40 correct trials were followed by reward feedback. Correct identification of the rich stimulus was associated with significantly more positive feedback (30 out of 40 trials) compared to correct identification of the lean stimulus (10 out of 40 trials). 
The EMBARC study was designed to discern differences in reward learning abilities between patients diagnosed with Major Depressive Disorder (MDD) and a healthy control group.
The PRT experiments involved 40 subjects from the control group and 168 subjects from the MDD group. Each subject in the MDD group had two PRT sessions, one session at the baseline before treatment (week 0) and one session after one week of treatment (week 1). Subjects in the control group also took two sessions in week 0 and week 1  (no treatment at both times). More details about the EMBARC study and the PRT experiment can be found in \cite{guo2023semiparametric}.

\vspace{-1.5em}
\subsection{Model fitting and results}

We now apply the proposed methodology to our motivating PRT data.
Let $\mathcal{S} = \{0, 1\}$, where $0$ and $1$ represent the type of stimulus (lean versus rich); $\mathcal{A} = \{0, 1\}$, where $0$ and $1$ represent the subject choosing lean and rich stimulus, respectively.
We choose the basis $\phi(a,s) = (\delta_{00}, \delta_{10}, \delta_{01}, \delta_{11})^{\top}$, where $\delta_{uv} = I(a=u, s=v)$ $u,v \in \{0,1\}$.
We assume no initial preference for one stimulus state over the other at the beginning of each session. To ensure this, we specify the initial coefficients as $\balpha = (\alpha/\rho) \cdot (1, 0, 0, 1)$ and set the RL intercept values as $\nu_0 = \nu_1 = 0$.
In a preliminary analysis, we observed that the learning patterns may change between the first and second block. To mitigate any potential bias, we analyze the PRT data from the first block in this paper. We started by fitting a separate RL model for each session, using the method outlined in Section \ref{sec:RL}, and excluded sessions with a learning rate less than $10^{-3}$.
We fitted our models for the MDD group that contains subjects diagnosed with MDD at week 0 (pre-treatment) and the control group with two repeated measurements, respectively. 
Based on the $5$-fold cross-validation likelihood score, we choose an RL-HMM with fixed transition probabilities over trials. This model has a higher score than an RL-HMM with time-varying  transition probabilities or an RL-only model without decision-making strategy switching. The transition probabilities, $C_{01}$, are $2.8\%$ for the MDD group and $1.3\%$ for the control group. Similarly, the transition probabilities, $C_{11}$, are $99.2\%$ for the MDD group and $99.7\%$ for the control group.
For inference, a nonparametric bootstrap was applied to generate 200 resampling sets for each group. Bootstrap standard errors and $95\%$ bootstrap confidence intervals are shown in Table \ref{table:pars_est}.

Table \ref{table:pars_est} reveals that the differences in the learning rate ($\beta$), reward sensitivity ($\rho$), and the initial probability of being in the `engaged' strategy ($\pi_1$) between the MDD group and the control group are not statistically significant. This finding contrasts with previous research, such as \cite{huys2013mapping} and \cite{guo2023semiparametric}, in which they reported a significantly lower reward sensitivity in the MDD group compared to the healthy control group in RL-only models. 
On the other hand, the MDD group presents a significantly lower initial coefficient ($\alpha$) compared to the control group. This indicates that the MDD group has a lower probability of selecting the correct state (i.e., distinguishing between long and short mouths) at the beginning of the task.

\begin{figure}[!p]
    \centering
        \begin{tabular}{cc}
          \includegraphics[scale=0.21]{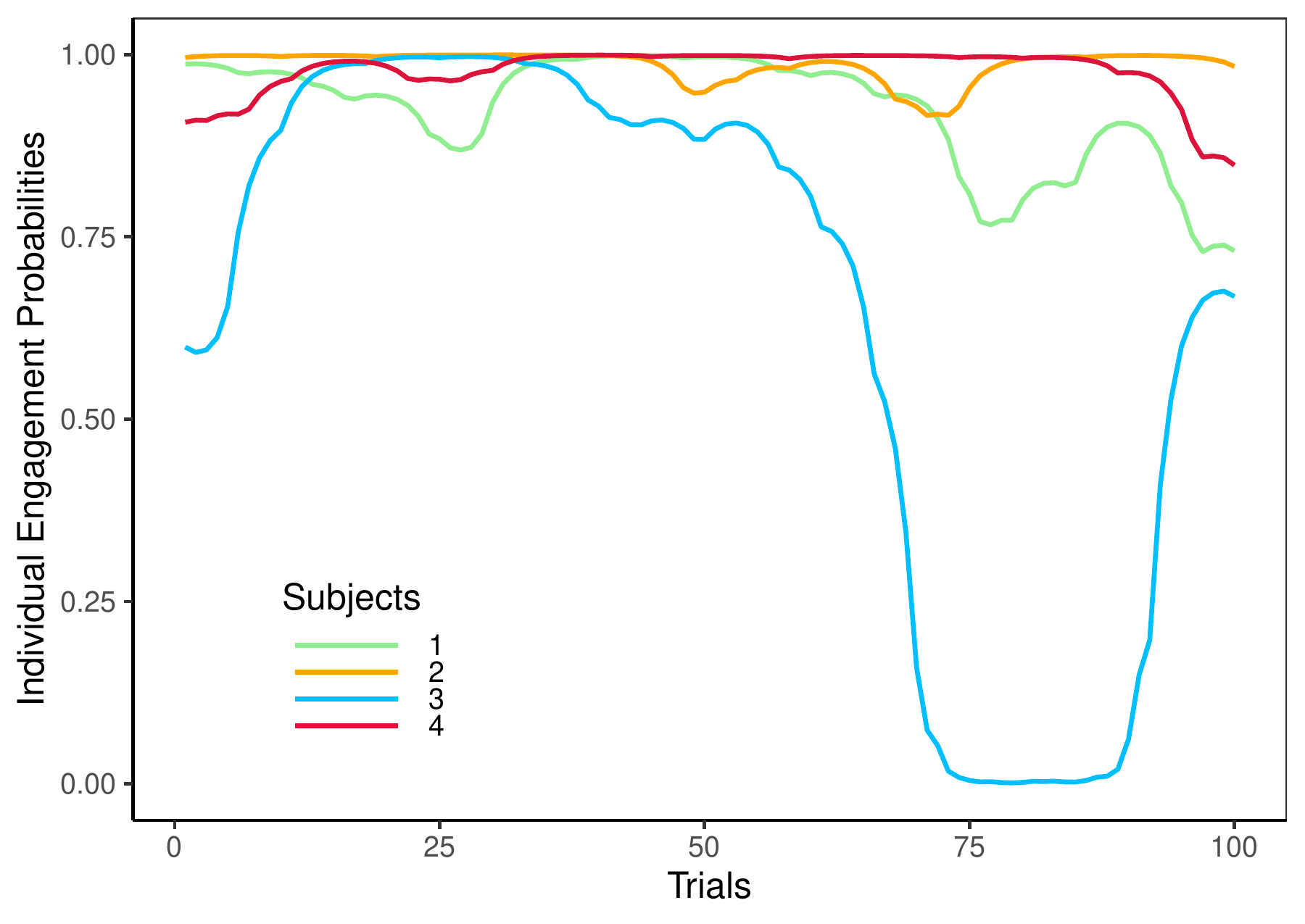} & 
          \includegraphics[scale=0.21]{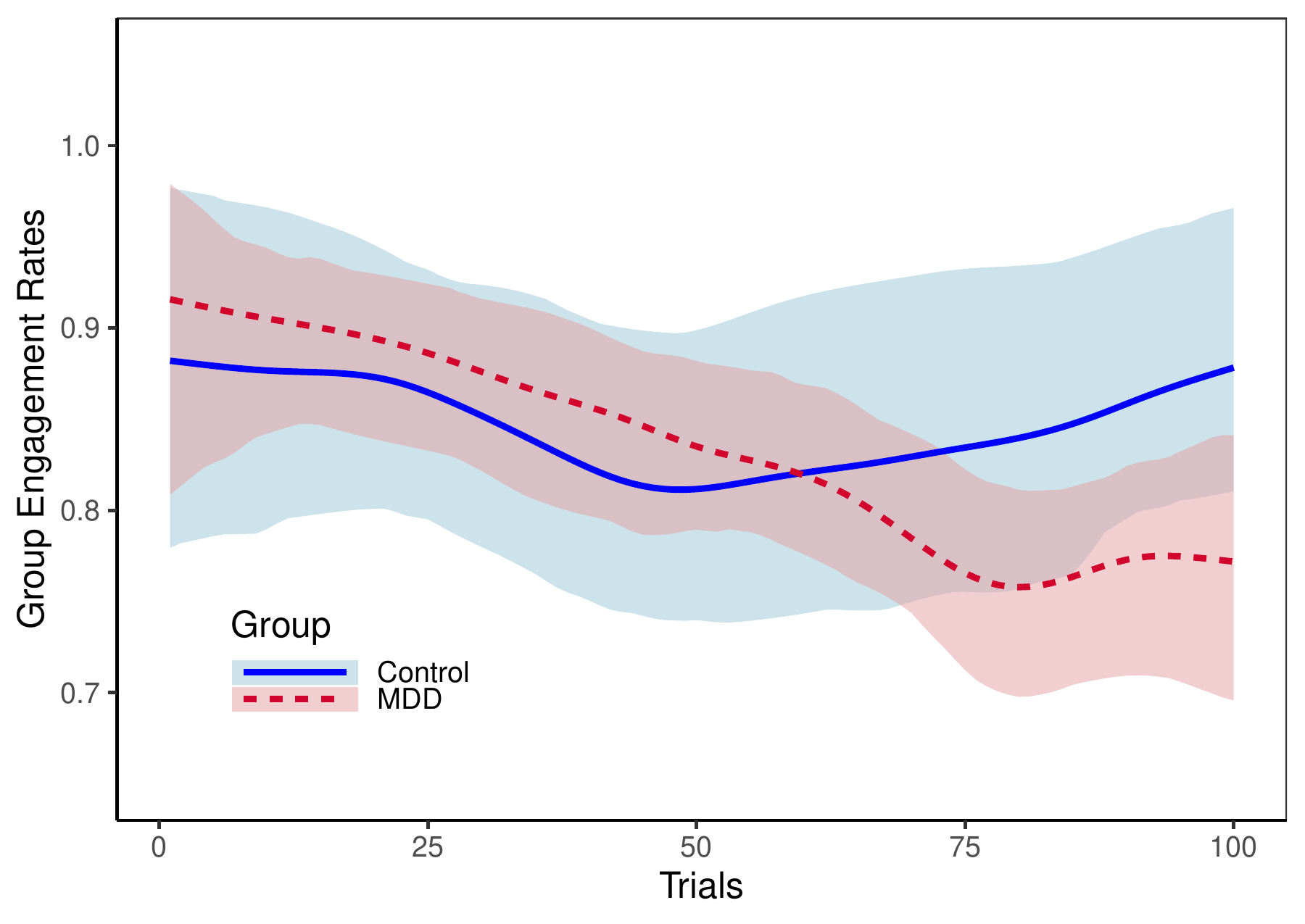} \\
          (a) Individual engagement probabilities & (b) Group engagement rates   \vspace{2em} \\
         
          \includegraphics[scale=0.21]{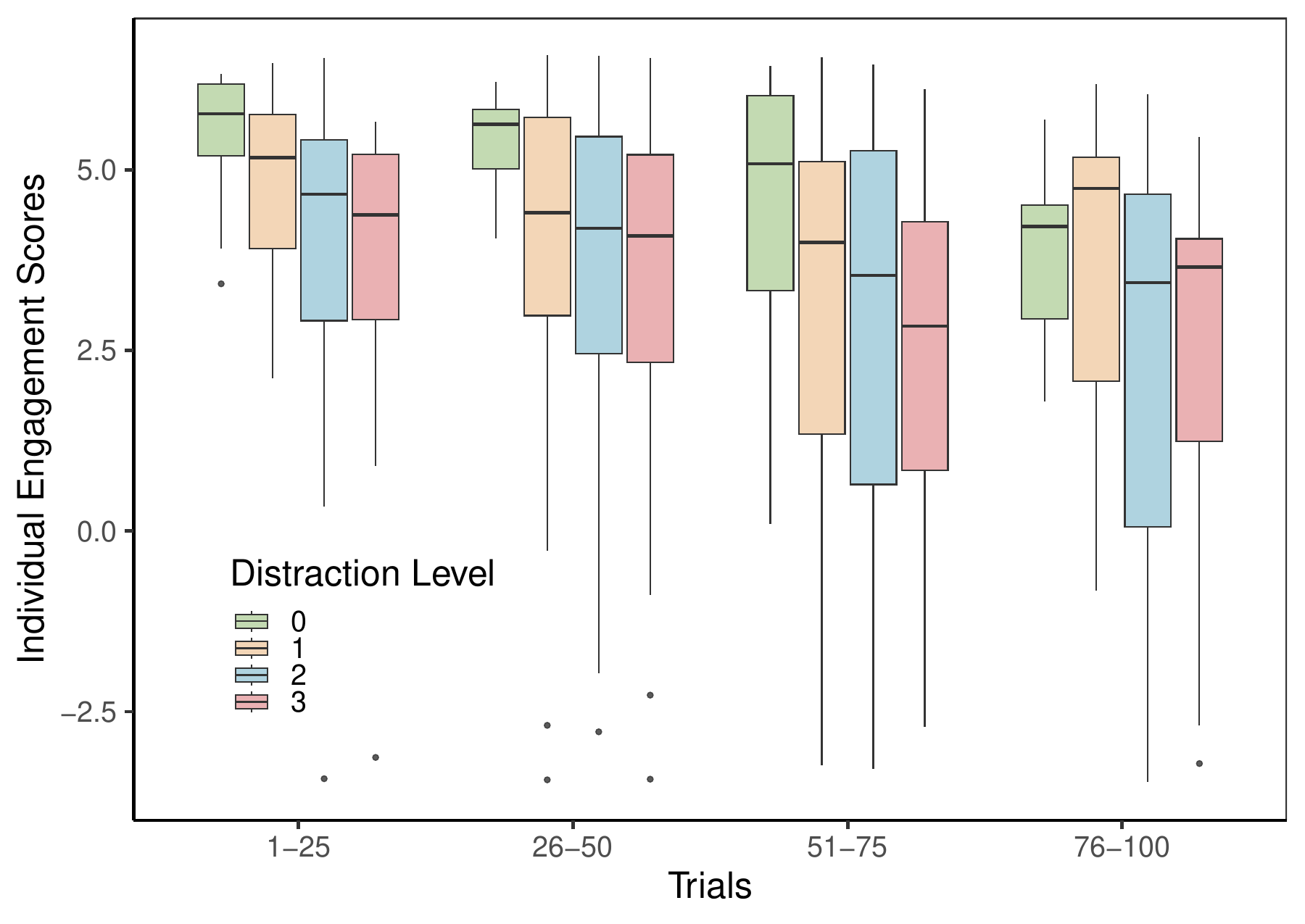} & 
          \includegraphics[scale=0.21]{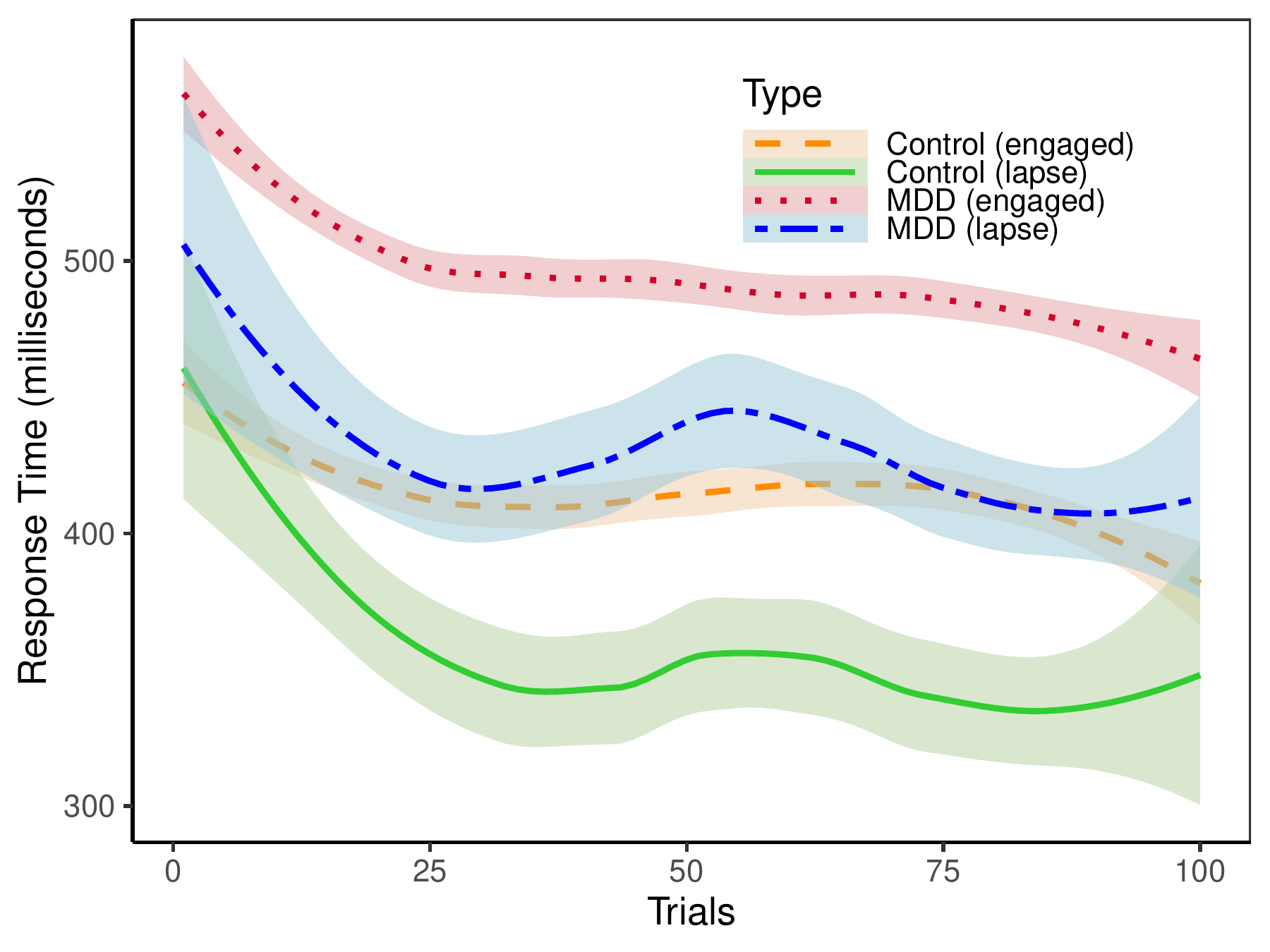} \\
          (c) Individual engagement scores & (d) Response time (ITIs)  \\
        \end{tabular}
    \caption{Estimation of individual engagement probabilities for four randomly selected MDD patients (Penal a); MDD/control group engagement rates (Penal b); comparison of individual engagement scores versus distraction levels (Penal c), and variation in response time (ITIs) across decision-making strategies and groups (Penal d).}
    \label{img:real}
\end{figure}

Figure \ref{img:real}(a) presents the plug-in estimates of the individual engagement probability, $\widehat{\gamma}_{i1t} = P(U_{it}=1 \mid \mathbf{A}_i, \widehat{\btheta} )$, for four randomly selected MDD patients which shows a general decreasing tendency of being engaged but with heterogeneous patterns between patients.
The estimates and the $95\%$ pointwise bootstrap confidence intervals for the MDD and control groups engagement rates (defined in Section \ref{sec:RL-HMM}) are shown in Figure \ref{img:real}(b). We observe that the engagement rates for MDD and control groups are similar in the first 3/4 of the task. However, the MDD group engagement rate decreases and becomes smaller in the last 1/4 of the task compared to the control group engagement rate. This implies that individuals in the MDD group are more likely to be in the `lapse' state after many trials in contrast to the control group. 
This finding suggests an alternative explanation for the reward processing abnormalities in MDD; it may be attributed to a reduced duration of making decisions using RL rather than a diminished sensitivity to rewards.

To demonstrate the utilities of individual engagement scores (defined in Section \ref{sec:RL-HMM}), we consider its association with a commonly used clinical measure of depression severity, the Hamilton Depression Rating Scale (HDRS) \citep{hamilton1960rating}. In EMBARC, the HDRS was administered by trained clinicians to assess depression symptoms in different domains at the same time when the PRT was collected. One of items in HDRS assesses concentration (referred to as distraction level) on a liker scale of zero to three, where a higher value indicates greater difficulty in concentrating.
In Figure \ref{img:real}(c), we present boxplots of individual engagement scores at different distraction levels (ranging from 0 to 3) in four periods of the task for patients in the MDD group. We observe that as the distraction level increases, the individual engagement score decreases. This effect is more pronounced in the first 3/4 of the trials.
Furthermore, we performed a regression analysis of distraction levels on the individual engagement score using generalized linear models with binomial link functions. We found that the $p$-values for the  coefficients are $0.004$, $0.043$, $0.013$, and $0.090$ for the four periods of the tasks, respectively. This implies that a lack of engagement, particularly at the beginning of the task, is associated with difficulties in concentration.

Define $\widehat{U}_{it} = I(\widehat{\gamma}_{i1t} \ge 0.5)$ to be the predicted type of decision-making strategy for the subject $i$ at trial $t$. 
The correct response rates (i.e. the accurate identification of short and long mouth stimuli) are $80.9\%$ and $48.4\%$ for the `engaged' and `lapse' state within the MDD group, and $84.3\%$ and $55.3\%$ for the control group. This suggests that the MDD group has a lower correct response rate in the `engaged' state, and the correct response rate in the `lapse' state is close to random guessing in both groups. 
Additionally, \cite{iigaya2018effect} showed that different decision-making strategies may result in different lengths of intertrial intervals (ITIs), which is the response time between the presence of a stimulus and the decision making in a trial.
Furthermore, we obtain the smoothed ITIs over trials for the two decision-making strategies for two groups separately using local polynomial regression (\textit{loess} function in R). 
The mean estimates and their $95\%$ confidence intervals for the mean curves are displayed in Figure \ref{img:real}(d). It reveals that, on average, the `engaged' strategy requires more time for individuals to make decisions compared to the `lapse' strategy. Meanwhile, the control group, on average, takes less time to make decisions than the MDD group.

\vspace{-1.5em}
\subsection{Brain-behavior Association}

We examine whether abnormalities in reward processing may reflect underlying changes in the brain circuits that regulate reward-related behavior and learning. Specifically,  we investigate whether dysfunctions in MDD-implicated brain circuitries affect learning abilities by an association analysis of subjects' individual engagement scores and task functional magnetic resonance imaging (fMRI) measures of brain activation. We focus on fMRI measures in an emotional conflict task assessing amygdala-anterior cingulate (ACC) circuitry \citep{etkin2006resolving,fonzo2019brain}. The emotional conflict task comprises four types of trials: congruent (C) trials, incongruent (I) trials, incongruent followed by incongruent (iI) trials, and incongruent followed by congruent (cI) trials.

To investigate the brain-behavior association for each brain region of interest (ROI),  the voxel-level activation conflicts (I-C) and activation conflict adaptations (iI - cI) within each ROI for each subject in the MDD group before treatment were computed. To investigate interregional interaction between ROIs, a psychophysiological interaction (PPI) analysis, as outlined in \cite{etkin2006resolving}, provides voxel-wise conflict adaptation coupling between two ROIs. As a result, we considered six types of fMRI measures: (i) mean of activation conflicts within ROIs (I-C mean), (ii) mean of activation conflict adaptations within ROIs (iI-cI mean), (iii) mean of conflict adaptation coupling between two ROIs (PPI mean), (iv) standard deviation of activation conflicts within ROIs (I-C std), (v) standard deviation of activation conflict adaptations within ROIs (iI-cI std), and (vi) standard deviation of conflict adaptation coupling between two ROIs (PPI std).
We utilize linear regressions to model the associations between individual engagement scores and each fMRI measure for all ROIs/interactions. Subsequently, we calculate the p-values for the regression coefficients and implement q-values for controlling the false discovery rate (FDR), as proposed by \citep{benjamini1995controlling}.

\begin{figure}[!p]
    \begin{subfigure}{0.54\textwidth}
        \centering
        \includegraphics[width=0.9\linewidth]{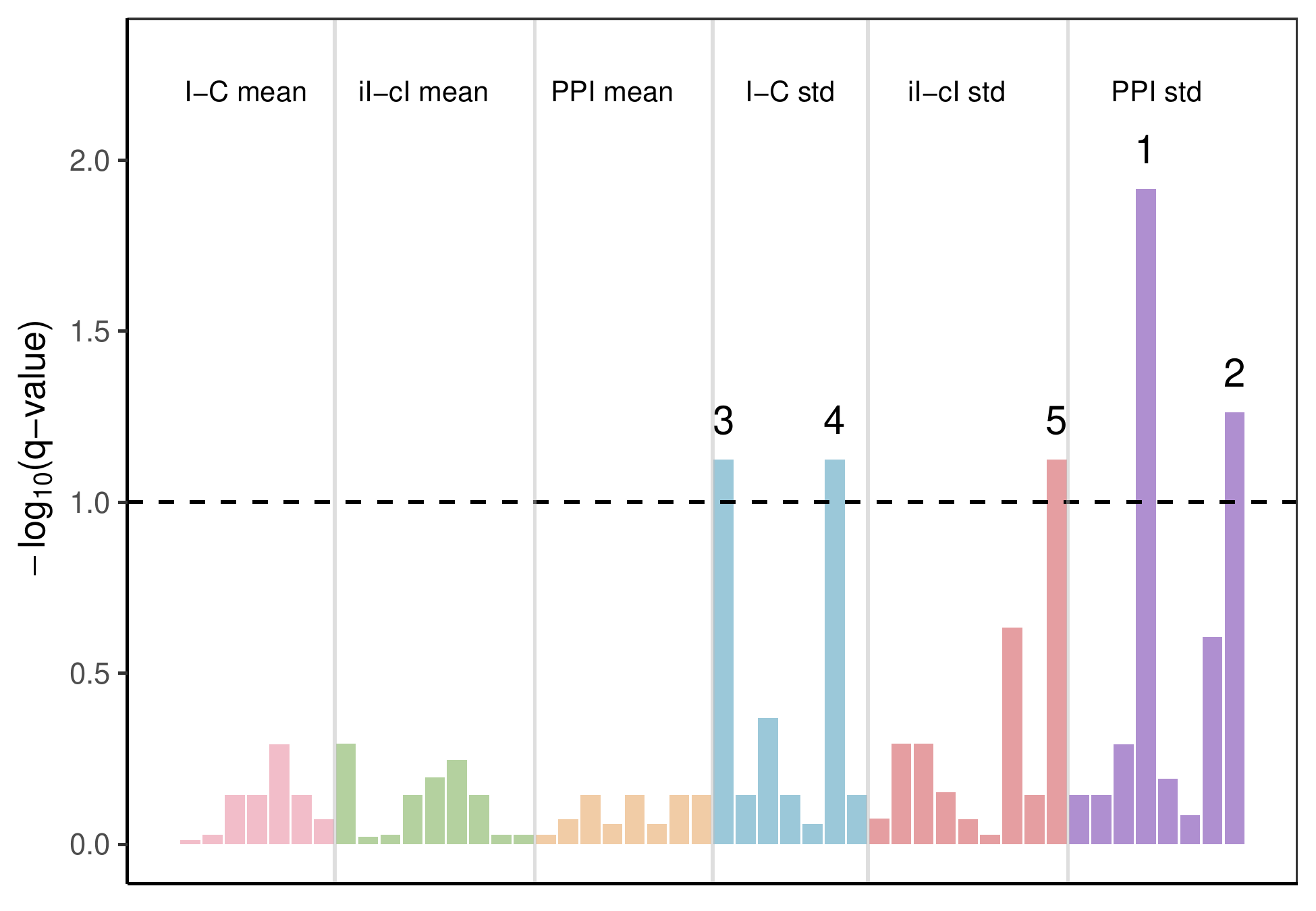}
        \caption{$-\log_{10}(\text{q-value})$ of the brain-behavior association}
    \end{subfigure}
    \begin{subfigure}{0.44\textwidth}
        \centering
        \includegraphics[width=0.90\linewidth]{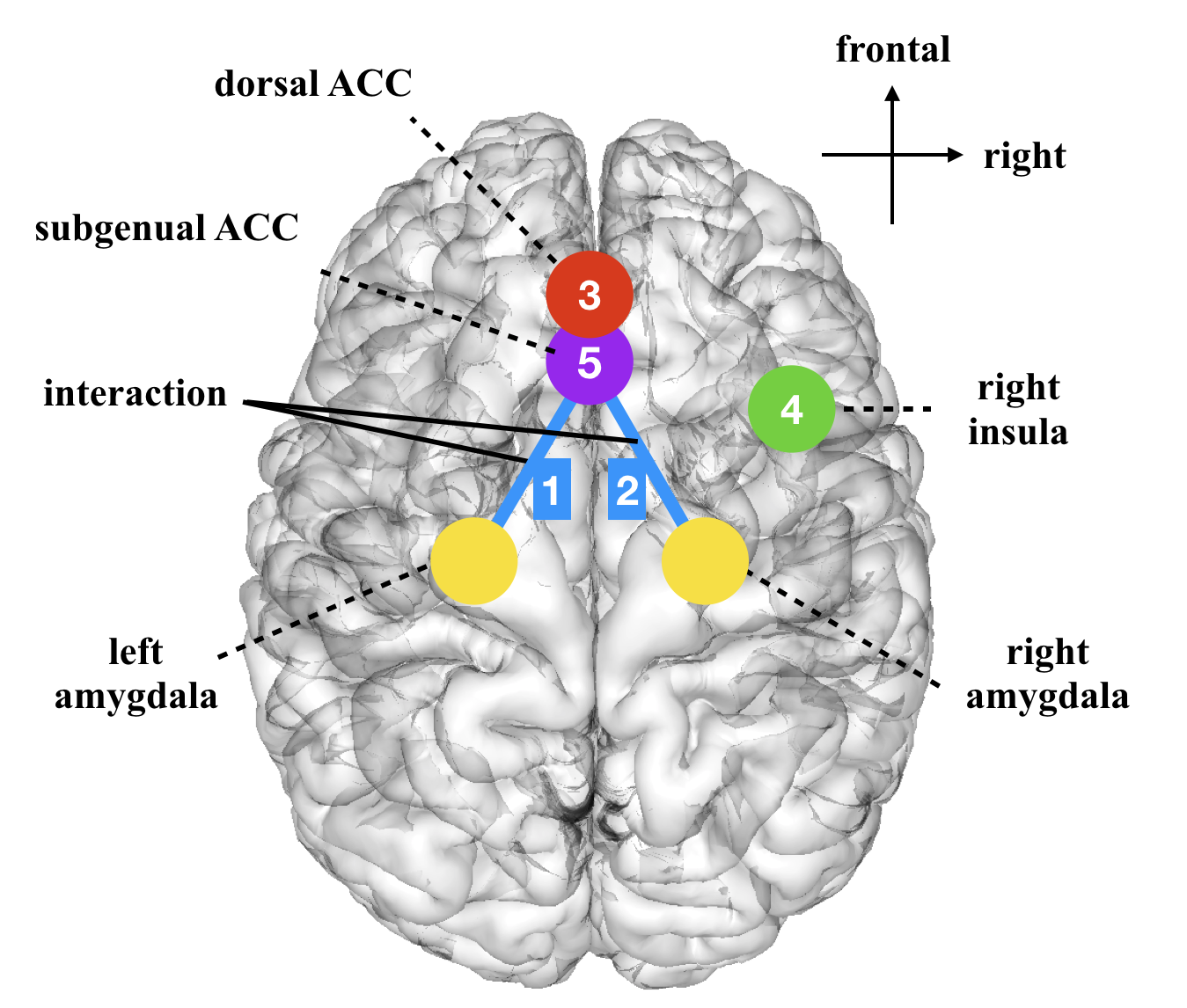}
        \caption{ROIs and ROI interactions identified}
    \end{subfigure}
    \caption{(a): The ($-\log_{10}$ transformation of) q-values of the regression coefficients for individual engagement scores regressed on each fMRI measure across all ROIs/interactions. The dashed line indicates the FDR at $10\%$. (b): Visualization of significant ROIs/interactions in the brain, with each number (1 to 5) corresponding to the respective q-value in (a).}
  	\label{img:fmri}
\end{figure}

The results are depicted in Figure \ref{img:fmri}. In Figure \ref{img:fmri}(a), each $q$-value undergoes a transformation using a negative $\log_{10}$ function, with the dashed line representing the FDR at $10\%$. Three ROIs and two ROI interactions were detected. A visualization illustrating the detected ROIs or interactions in the brain is presented in Figure \ref{img:fmri}(b), where the associations between engagement scores and the brain response to emotional conflict adaptation coupling of subgenual ACC ROI and left amygdala ROI appears to be the most significant. Other significant associations include the conflict coupling of subgenual ACC and right amygdala, the conflict within dorsal ACC and right insula, and the conflict adaptations within the subgenual ACC.
In this analysis, all significant brain measures identified were the standard deviations of ROIs, rather than their mean effects. In addition, our results show a negative correlation between engagement scores and the standard deviations of all five identified ROIs or interactions. This suggests that an increased engagement in reward learning tasks corresponds to a decreased variability in brain activity during an emotional conflict task.
\cite{williams2016precision} further indicated that the ACC---amygdala connectivity belongs to the negative affect circuit. Hence, MDD patients are more prone to lose concentration during the reward learning behavioral task, which could be due to an increase of fluctuation (or instability) of  brain activities within the negative affect circuit.

\vspace{-1.5em}
\section{Discussion}
\label{sec:discussion}
\vspace{-0.75em}
In this paper, we propose a RL-HMM method to characterize reward-based decision-making, incorporating the switching of decision-making strategies between two states: the `engaged' state, where subjects make decisions based on the RL model, and the `lapse' state, where they make random choices. We extend the RL model to accommodate a non-discrete state space and posit time-varying transition probabilities within the HMM.
For parameter estimation, we employ the EM algorithm. In the E-step, we apply forward-backward inductive computation, and in the M-step, we implement a computationally efficient algorithm for one-step updates of all the parameters. Extensive simulation studies validate the finite-sample performance of our method. 

Applying our approach to the EMBARC, we discovered different phenomena from the findings in \cite{huys2013mapping} and \cite{guo2023semiparametric}, where the MDD group had a lower reward sensitivity compared to the healthy controls. In contrast, our method did not identify a significant group difference in reward sensitivity. Instead, we discovered that the MDD group spent a larger proportion of time using the `lapse' strategy compared to the healthy group. This suggests an alternative mechanism of the reward processing abnormalities in MDD.
We identified a significant correlation between the individual engagement score and a clinician-assessed measure of concentration difficulty. This finding suggests that  tendencies towards disengagement may be linked with a loss of concentration ability in MDD.
Moreover, our analysis revealed that the `lapse' strategy requires less time for individuals to respond than the `engaged' strategy in PRT.
Using additional brain fMRI data within the same study, we found that the individual engagement scores computed from RL-HMM are associated with the variability of brain activities in the negative affect circuitry under an emotional conflict task \citep{etkin2006resolving}.
This brain-behavior association study suggests that a higher heterogeneity in brain activities during emotional conflict is associated with lower engagement in reward learning tasks.

In PRT, The individual engagement score serves as a crucial behavioral marker. However, our current work does not accommodate heterogeneity beyond the aspects of the pattern of individual engagement probabilities during tasks. Some extensions include introducing covariates to model between-subjects heterogeneity in RL parameters (such as learning rate, reward sensitivity, and initial coefficients of the expected reward) and transition probabilities in the HMM. 
Our methods are sufficiently general to analyze other types of behavioral tasks. 
Another extension involves jointly modeling PRT with other phenotypes, including brain measures and clinical outcomes.

Recall that we assume subjects make decisions independently of how they learn from rewards. 
In psychology, humans may learn from rewards without conscious awareness; this phenomenon is termed implicit learning \citep{frensch2003implicit}. It suggests that individuals can unconsciously detect patterns and learn from the outcomes of their actions over time.
In the reinforcement learning (RL) literature, another phenomenon, exploration versus exploitation, may help in understanding our assumption: Exploration in RL involves testing new actions to discover potential rewards, while exploitation entails selecting actions known to yield high rewards. Even within an exploration strategy, reward learning still occurs \citep{sutton2018reinforcement}.
A generalization of our proposed model is that we assume reward learning occurs in different decision-making strategies, and the learning behavior—encoded by RL parameters such as learning rate and reward sensitivity—might vary depending on the strategy. For example, individuals in the `lapse' state may learn more slowly or be less sensitive to reward.



\vspace{-2em}
\section{Supporting Materials}
\vspace{-0.8em}
\label{sec6}
Supplementary materials containing additional algorithm and simulation details are available online at \url{http://biostatistics.oxfordjournals.org}.  The R code for the simulation studies is available on GitHub at \url{https://github.com/xingcheg/RL-HMM}.  

\vspace{-1.5em}
\section*{Acknowledgments}
\vspace{-0.8em}
This research is supported by U.S. NIH grants MH123487, NS073671 and GM124104.


\begin{thebibliography}{99}
\vspace{-1em}
\bibitem[Abbeel and Ng, 2004]{abbeel2004apprenticeship}
Abbeel, Pieter and Ng, Andrew Y. (2004). Apprenticeship Learning via Inverse Reinforcement Learning. In: \emph{Proceedings of the Twenty-first International Conference on Machine Learning}.

\vspace{-0.5em}
\bibitem[Arnold and Tibshirani, 2016]{arnold2016efficient}
Arnold, Taylor B and Tibshirani, Ryan J. (2016). Efficient implementations of the generalized lasso dual path algorithm. \emph{Journal of Computational and Graphical Statistics} 25(1), 1--27.

\vspace{-0.5em}
\bibitem[Ashwood et al., 2022]{ashwood2022mice}
Ashwood, Zoe C, Roy, Nicholas A, Stone, Iris R, Urai, Anne E, Churchland, Anne K, Pouget, Alexandre and Pillow, Jonathan W.  (2022). Mice alternate between discrete strategies during perceptual decision-making. \emph{Nature Neuroscience} 25(2), 201--212.

\vspace{-0.5em}
\bibitem[Baum et al., 1970]{baum1970maximization}
Baum, Leonard E, Petrie, Ted, Soules, George and Weiss, Norman. (1970). A maximization technique occurring in the statistical analysis of probabilistic functions of Markov chains. \emph{The Annals of Mathematical Statistics} 41(1), 164--171.

\vspace{-0.5em}
\bibitem[Benjamini and Hochberg, 1995]{benjamini1995controlling}
Benjamini, Yoav and Hochberg, Yosef. (1995). Controlling the false discovery rate: a practical and powerful approach to multiple testing. \emph{Journal of the Royal Statistical Society: Series B (Methodological)} 57(1), 289--300.

\vspace{-0.5em}
\bibitem[Byrd et al., 1995]{byrd1995limited}
Byrd, Richard H, Lu, Peihuang, Nocedal, Jorge and Zhu, Ciyou. (1995). A limited memory algorithm for bound constrained optimization. \emph{SIAM Journal on Scientific Computing} 16(5), 1190--1208.

\vspace{-0.5em}
\bibitem[Chen et al., 2021]{chen2021sex}
Chen, Cathy S, Knep, Evan, Han, Autumn, Ebitz, R Becket and Grissom, Nicola M. (2021). Sex differences in learning from exploration. \emph{Elife} 10, e69748.

\vspace{-0.5em}
\bibitem[Dempster et al., 1977]{dempster1977maximum}
Dempster, Arthur P, Laird, Nan M and Rubin, Donald B. (1977). Maximum likelihood from incomplete data via the EM algorithm. \emph{Journal of the Royal Statistical Society: Series B (Statistical Methodology)} 39(1), 1--22.

\vspace{-0.5em}
\bibitem[Etkin et al., 2006]{etkin2006resolving}
Etkin, Amit, Egner, Tobias, Peraza, Daniel M, Kandel, Eric R and Hirsch, Joy. (2006). Resolving emotional conflict: A role for the rostral anterior cingulate cortex in modulating activity in the amygdala. \emph{Neuron} 51(6), 871--882.

\vspace{-0.5em}
\bibitem[Fonzo et al., 2019]{fonzo2019brain}
Fonzo, Gregory A, Etkin, Amit, Zhang, Yu, Wu, Wei, Cooper, Crystal, Chin-Fatt, Cherise, Jha, Manish K, Trombello, Joseph, Deckersbach, Thilo, Adams, Phil et al. (2019). Brain regulation of emotional conflict predicts antidepressant treatment response for depression. \emph{Nature Human Behaviour} 3(12), 1319--1331.

\vspace{-0.5em}
\bibitem[Frensch and R{\"u}nger, 2003]{frensch2003implicit}
Frensch, Peter A and R{\"u}nger, Dennis. (2003). Implicit learning. \emph{Current Directions in Psychological Science} 12(1), 13--18.

\vspace{-0.5em}
\bibitem[Guo et al., 2024]{guo2023semiparametric}
Guo, Xingche, Zeng, Donglin and Wang, Yuanjia. (2024). A semiparametric inverse reinforcement learning approach to characterize decision making for mental disorders. \emph{Journal of the American Statistical Association} 119(545), 27--38.

\vspace{-0.5em}
\bibitem[Hamilton, 1960]{hamilton1960rating}
Hamilton, Max. (1960). A rating scale for depression. \emph{Journal of Neurology, Neurosurgery, and Psychiatry} 23(1), 56.

\vspace{-0.5em}
\bibitem[Huys et al., 2011]{huys2011disentangling}
Huys, Quentin JM, Cools, Roshan, G\"{o}lzer, Martin, Friedel, Eva, Heinz, Andreas, Dolan, Raymond J and Dayan, Peter. (2011). Disentangling the roles of approach, activation and valence in instrumental and pavlovian responding. \emph{PLoS Computational Biology} 7(4), e1002028.

\vspace{-0.5em}
\bibitem[Huys et al., 2016]{huys2016computational}
Huys, Quentin JM, Maia, Tiago V and Frank, Michael J. (2016). Computational psychiatry as a bridge from neuroscience to clinical applications. \emph{Nature Neuroscience} 19(3), 404--413.

\vspace{-0.5em}
\bibitem[Huys et al., 2013]{huys2013mapping}
Huys, Quentin JM, Pizzagalli, Diego A, Bogdan, Ryan and Dayan, Peter.  (2013). Mapping anhedonia onto reinforcement learning: A behavioural meta-analysis. \emph{Biology of Mood \& Anxiety Disorders} 3(1), 1--16.

\vspace{-0.5em}
\bibitem[Iigaya et al., 2018]{iigaya2018effect}
Iigaya, Kiyohito, Fonseca, Madalena S, Murakami, Masayoshi, Mainen, Zachary F and Dayan, Peter. (2018). An effect of serotonergic stimulation on learning rates for rewards apparent after long intertrial intervals. \emph{Nature Communications} 9(1), 1--10.

\vspace{-0.5em}
\bibitem[Insel et al., 2010]{insel2010research}
Insel, Thomas, Cuthbert, Bruce, Garvey, Marjorie, Heinssen, Robert, Pine,
Daniel S, Quinn, Kevin, Sanislow, Charles and Wang, Philip.  (2010). Research domain criteria (RDoC): Toward a new classification framework for research on mental disorders. \emph{American Journal of Psychiatry} 167(7), 748--751.

\vspace{-0.5em}
\bibitem[Kendler et al., 1999]{kendler1999causal}
Kendler, Kenneth S, Karkowski, Laura M and Prescott, Carol A. (1999). Causal relationship between stressful life events and the onset of major depression. \emph{American Journal of Psychiatry}, 156(6), 837--841.

\vspace{-0.5em}
\bibitem[Pizzagalli et al., 2005]{pizzagalli2005toward}
Pizzagalli, Diego A, Jahn, Allison L and O’Shea, James P. (2005). Toward an objective characterization of an anhedonic phenotype: A signal-detection approach. \emph{Biological Psychiatry}, 57(4), 319--327.

\vspace{-0.5em}
\bibitem[Rescorla, 1972]{rescorla1972theory}
Rescorla, Robert A. (1972). A theory of Pavlovian conditioning: Variations in the effectiveness of reinforcement and nonreinforcement. \emph{Current Research and Theory}, 64--99.

\vspace{-0.5em}
\bibitem[Ross and Bagnell, 2010]{ross2010efficient}
Ross, St\'ephane and Bagnell, Drew. (2010). Efficient reductions for imitation learning. In: \emph{Proceedings of the Thirteenth International Conference on Artificial Intelligence and Statistics}, JMLR Workshop and Conference Proceedings, pp. 661--668.

\vspace{-0.5em}
\bibitem[Rush et al., 2006]{rush2006report}
Rush, A John, Kraemer, Helena C, Sackeim, Harold A, Fava, Maurizio, Trivedi, Madhukar H, Frank, Ellen, Ninan, Philip T, Thase, Michael E, Gelenberg, Alan J, Kupfer, David J et al. (2006). Report by the ACNP task force on response and remission in major depressive disorder. \emph{Neuropsychopharmacology}, 31(9), 1841--1853.

\vspace{-0.5em}
\bibitem[Schultz et al., 1997]{schultz1997neural}
Schultz, Wolfram, Dayan, Peter and Montague, P Read.  (1997). A neural substrate of prediction and reward. \emph{Science}, 275(5306), 1593--1599.

\vspace{-0.5em}
\bibitem[Sutton and Barto, 2018]{sutton2018reinforcement}
Sutton, Richard S and Barto, Andrew G. (2018). \emph{Reinforcement Learning: An Introduction}. MIT Press.


\vspace{-0.5em}\bibitem[Tibshirani et al., 2005]{tibshirani2005sparsity}
Tibshirani, Robert, Saunders, Michael, Rosset, Saharon, Zhu, Ji and Knight,
Keith. (2005). Sparsity and smoothness via the fused lasso. \emph{Journal of the Royal Statistical Society: Series B (Statistical Methodology)}, 67(1), 91--108.

\vspace{-0.5em}
\bibitem[Tibshirani, 2014]{tibshirani2014adaptive}
Tibshirani, Ryan J.  (2014). Adaptive piecewise polynomial estimation via trend filtering. \emph{The Annals of Statistics}, 42(1), 285--323.

\vspace{-0.5em}
\bibitem[Tibshirani and Taylor, 2011]{tibshirani2011solution}
Tibshirani, Ryan J and Taylor, Jonathan. (2011). The solution path of the generalized lasso. \emph{Annals of Statistics}, 39(3), 1335--1371.

\vspace{-0.5em}
\bibitem[Torabi et al., 2018]{torabi2018behavioral}
Torabi, Faraz, Warnell, Garrett and Stone, Peter. (2018). Behavioral cloning from observation. \emph{arXiv preprint arXiv:1805.01954}.

\vspace{-0.5em}
\bibitem[Trivedi et al., 2016]{trivedi2016establishing}
Trivedi, Madhukar H, McGrath, Patrick J, Fava, Maurizio, Parsey, Ramin V, Kurian, Benji T, Phillips, Mary L, Oquendo, Maria A, Bruder, Gerard, Piz- zagalli, Diego, Toups, Marisa et al. (2016). Establishing moderators and biosignatures of antidepressant response in clinical care (EMBARC): Rationale and design. \emph{Journal of Psychiatric Research}, 78, 11--23.

\vspace{-0.5em}
\bibitem[Williams, 2016]{williams2016precision}
Williams, Leanne M. (2016). Precision psychiatry: A neural circuit taxonomy for depression and anxiety. \emph{The Lancet Psychiatry}, 3(5), 472--480.

\vspace{-0.5em}
\bibitem[Worthy et al., 2013]{worthy2013heterogeneity}
Worthy, Darrell A, Hawthorne, Melissa J and Otto, A Ross. (2013). Heterogeneity of strategy use in the Iowa Gambling Task: A comparison of win-stay/lose-shift and reinforcement learning models. \emph{Psychonomic Bulletin \& Review}, 20(2), 364--371.

\end{thebibliography}

\vspace{-2em}

\bibliographystyle{natbib}

\begin{table}[!p]
\caption{Summary of the parameter estimates in 200 simulations for Case I (data generated from RL-HMM).}
\begin{center}\vskip -.15in
\footnotesize
\begin{tabular}{lllrrrrrrrrrr}
\toprule
 & & & \multicolumn{4}{c}{RL-HMM} && \multicolumn{2}{c}{RL-HMM-fixed} && \multicolumn{2}{c}{RL-only} \\
 \midrule
$T$   &   $n$ &    Parameters     & Bias  & SD          & SE       & CP  && Bias  & SD  && Bias  & SD \\
\cline{4-7}\cline{9-10} \cline{12-13}\\
100 & 100 & $\beta$               & -0.0005  & 0.0073   & 0.0076   & 93  && -0.0067 & 0.0070 && -0.0070 & 0.0166 \\
    &     & $\rho$                &  0.1004  & 0.3741   & 0.3664   & 94  &&  0.0775 & 0.3674 && -1.7117 & 0.3954 \\
    &     & $\nu_1$               & -0.0073  & 0.0790   & 0.0755   & 93  && -0.0357 & 0.0708 && -0.0003 & 0.0539 \\
    &     & $\alpha$              &  0.0182  & 0.1133   & 0.1211   & 98  && -0.1577 & 0.1085 && -0.9368 & 0.0571 \\
    &     & $\pi_1$               & -0.0148  & 0.1346   & 0.1269   & 94  && -0.0460 & 0.1030 \\
    &     & $\zeta_0(T/4)$        & -0.1326  & 0.2693   & 0.2908   & 93  && -1.0037 & 0.1787 \\
    &     & $\zeta_0(3T/4)$       &  0.1054  & 0.2575   & 0.2966   & 97  &&  0.4963 & 0.1787 &&\multicolumn{2}{c}{------------}\\
    &     & $\zeta_1(T/4)$        &  0.1111  & 0.2583   & 0.2816   & 96  &&  1.2878 & 0.2042 \\
    &     & $\zeta_1(3T/4)$       & -0.1410  & 0.2731   & 0.3158   & 94  && -0.7122 & 0.2042 \\
100 & 200 & $\beta$               & -0.0001  & 0.0054   & 0.0053   & 94  && -0.0066 & 0.0051 && -0.0079 & 0.0123 \\
    &     & $\rho$                &  0.0685  & 0.2594   & 0.2542   & 93  &&  0.0627 & 0.2590 && -1.7494 & 0.2923 \\
    &     & $\nu_1$               & -0.0023  & 0.0513   & 0.0527   & 97  && -0.0303 & 0.0468 && -0.0007 & 0.0354 \\
    &     & $\alpha$              &  0.0172  & 0.0832   & 0.0848   & 96  && -0.1494 & 0.0830 && -0.9404 & 0.0395 \\
    &     & $\pi_1$               & -0.0008  & 0.0942   & 0.0916   & 95  && -0.0357 & 0.0798 \\
    &     & $\zeta_0(T/4)$        & -0.1182  & 0.2044   & 0.2133   & 93  && -1.0036 & 0.1288 \\
    &     & $\zeta_0(3T/4)$       &  0.1022  & 0.1878   & 0.2013   & 96  &&  0.4964 & 0.1288 &&\multicolumn{2}{c}{------------}\\
    &     & $\zeta_1(T/4)$        &  0.0896  & 0.1966   & 0.2079   & 92  &&  1.2645 & 0.1697 \\
    &     & $\zeta_1(3T/4)$       & -0.1301  & 0.2134   & 0.2252   & 93  && -0.7355 & 0.1697 \\
200 & 100 & $\beta$               &  0.0000  & 0.0049   & 0.0050   & 93  && -0.0057 & 0.0047 && -0.0015 & 0.0108 \\
    &     & $\rho$                &  0.0729  & 0.2747   & 0.2661   & 93  &&  0.0584 & 0.2647 && -2.0046 & 0.2450 \\
    &     & $\nu_1$               & -0.0087  & 0.0655   & 0.0608   & 94  && -0.0561 & 0.0589 &&  0.0230 & 0.0480 \\
    &     & $\alpha$              &  0.0071  & 0.0948   & 0.0903   & 95  && -0.1587 & 0.0906 && -0.9270 & 0.0469 \\
    &     & $\pi_1$               & -0.0107  & 0.1333   & 0.1221   & 94  && -0.0377 & 0.1182 \\
    &     & $\zeta_0(T/4)$        & -0.0936  & 0.1840   & 0.1982   & 93  && -0.9678 & 0.1298 \\
    &     & $\zeta_0(3T/4)$       &  0.1125  & 0.1898   & 0.1974   & 91  &&  0.5322 & 0.1298 &&\multicolumn{2}{c}{------------}\\
    &     & $\zeta_1(T/4)$        &  0.1035  & 0.1706   & 0.1935   & 94  &&  1.2598 & 0.1505 \\
    &     & $\zeta_1(3T/4)$       & -0.1158  & 0.1854   & 0.2038   & 93  && -0.7402 & 0.1505 \\
200 & 200 & $\beta$               & -0.0004  & 0.0035   & 0.0035   & 95  && -0.0062 & 0.0033 && -0.0013 & 0.0074 \\
    &     & $\rho$                &  0.0532  & 0.1790   & 0.1863   & 94  &&  0.0511 & 0.1747 && -2.0268 & 0.1701 \\
    &     & $\nu_1$               & -0.0076  & 0.0434   & 0.0434   & 95  && -0.0548 & 0.0402 &&  0.0263 & 0.0339 \\
    &     & $\alpha$              &  0.0046  & 0.0641   & 0.0631   & 95  && -0.1543 & 0.0594 && -0.9235 & 0.0347 \\
    &     & $\pi_1$               &  0.0015  & 0.0913   & 0.0887   & 97  && -0.0376 & 0.0753 \\
    &     & $\zeta_0(T/4)$        & -0.0625  & 0.1447   & 0.1484   & 96  && -0.9533 & 0.0805 \\
    &     & $\zeta_0(3T/4)$       &  0.0652  & 0.1236   & 0.1424   & 95  &&  0.5467 & 0.0805 &&\multicolumn{2}{c}{------------}\\
    &     & $\zeta_1(T/4)$        &  0.0669  & 0.1366   & 0.1427   & 94  &&  1.2436 & 0.0940 \\
    &     & $\zeta_1(3T/4)$       & -0.0793  & 0.1352   & 0.1447   & 93  && -0.7564 & 0.0940 \\
\bottomrule
\end{tabular}
\end{center}
\label{table:1}
\footnotesize \textcolor{black}{(Bias): estimate bias; (SD): standard deviation; (SE): average bootstrap standard error; (CP): coverage probability of the $95\%$ bootstrap confidence intervals; (RL-HMM): our proposed method; (RL-HMM-fixed): RL-HMM model with time-invariant transition probabilities; (RL-only): the RL model without decision-making strategy switching.}
\end{table}

\begin{table}[!p]
\caption{\textcolor{black}{Parameter estimation in EMBARC study under the proposed method for MDD group, control group, and group contrast.}}
\centering
\small
\begin{tabular}{cccccccccc}
\toprule
 & \multicolumn{3}{c}{MDD} & & \multicolumn{3}{c}{Control} & & MDD - Control  \\
 Pars.  & Est.  & Err.  &  $95\%$ CI     & & Est.  & Err. &  $95\%$ CI &&   $95\%$ CI   \\
 \midrule
 $\beta$  & 0.031 & 0.009 & (0.012, 0.049) && 0.031 & 0.009 & (0.015, 0.048) && (-0.026, 0.025) \\
 $\rho$   & 3.338 & 0.327 & (2.696, 3.980) && 3.292 & 0.337 & (2.631, 3.953) && (-0.853, 0.945) \\
 $\alpha$ & 1.328 & 0.072 & (1.187, 1.470) && 1.853 & 0.177 & (1.506, 2.200) && \textbf{(-0.893, -0.155)}\\
 $\pi_1$  & 0.917 & 0.042 & (0.773, 0.973) && 0.882 & 0.057 & (0.659, 0.967) && (-0.985, 0.159) \\
\bottomrule
\end{tabular}
\label{table:pars_est}
\end{table}

\end{document}